\documentclass[11pt]{article}
\usepackage[a4paper, total={6.5in, 9.5in}]{geometry}
\usepackage[utf8]{inputenc} 
\usepackage[english]{babel}
\usepackage{booktabs}
\usepackage{lipsum}

\usepackage[dvipsnames]{xcolor}
\usepackage[colorlinks=true,linkcolor=blue]{hyperref}
\usepackage{url}            

\usepackage{mathtools}
\usepackage{amsmath}
\usepackage{amssymb}
\usepackage{amsfonts}      
\usepackage{amsthm}
\usepackage{nicefrac}       
\usepackage{microtype}
\usepackage[bb=boondox]{mathalfa}

\usepackage{algorithm}
\usepackage{algpseudocode}

\usepackage{siunitx}
\usepackage{graphicx}
\graphicspath{{./figures/}}
\DeclareGraphicsExtensions{.pdf}
\usepackage{subcaption}

\usepackage[capitalize]{cleveref}
\usepackage[numbers]{natbib}
\usepackage{thmtools, thm-restate}
\newtheorem{theorem}{Theorem}[section]

\newtheorem{assumption}{Assumption}

\newtheorem{remark}[theorem]{Remark}

\def\T{{\mathcal T}}
\def\X{{\mathcal X}}
\def\Y{{\mathcal Y}}
\def\G{{\mathcal G}}
\def\E{{\mathbb E}}
\def\f{{\hat f}}

\DeclareMathOperator*{\argmin}{arg\,min}

\title{Fast Hierarchical Games for Image Explanations}

\author{
  Jacopo~Teneggi\\
  Department of Biomedical Engineering\\
  Jonhs Hopkins University\\
  Baltimore, USA\\
  \texttt{jtenegg1@jhu.edu}\\
  \and
  Alexandre~Luster\\
  School of Life Sciences\\
  Ecole Polytechnique Fédérale de Lausanne\\
  CH-1015 Lausanne\\
  \texttt{alexandre.luster@epfl.ch}
  \and
  Jeremias~Sulam\\
  Department of Biomedical Engineering\\
  Jonhs Hopkins University\\
  Baltimore, USA\\
  \texttt{jsulam1@jhu.edu}\\
}

\begin{document}
\maketitle

\begin{abstract}
    As modern complex neural networks keep breaking records and solving harder problems, their predictions also become less and less intelligible. The current lack of interpretability often undermines the deployment of accurate machine learning tools in sensitive settings. In this work, we present a model-agnostic explanation method for image classification based on a hierarchical extension of Shapley coefficients--\emph{Hierarchical Shap (h-Shap)}--that resolves some of the limitations of current approaches. Unlike other Shapley-based explanation methods, h-Shap is scalable and can be computed without the need of approximation. Under certain distributional assumptions, such as those common in multiple instance learning, h-Shap retrieves the exact Shapley coefficients with an exponential improvement in computational complexity. We compare our hierarchical approach with popular Shapley-based and non-Shapley-based methods on a synthetic dataset, a medical imaging scenario, and a general computer vision problem, showing that h-Shap outperforms the state of the art in both accuracy and runtime. Code and experiments are made publicly available.
\end{abstract}

\section{Introduction}\label{sec:introduction}
Explainability has become a question of increasing relevance in machine learning, where the growing complexity of deep neural networks often renders them \emph{opaque} to us, the humans interacting with them. This issue is commonly referred to as the \emph{black-box problem} and comprises theoretical, technical, and regulatory questions \cite{Zednik2019SolvingIntelligence, Tomsett2018InterpretableSystems}. As deep neural networks take on sensitive tasks in medical, legal, and financial settings, they need to achieve both high accuracy and high transparency for a safe deployment. For example, uninterpretable predictions could mislead clinicians in their decision making rather than support it \cite{Amann2020ExplainabilityPerspective}. Furthermore, it is sometimes required by law \cite{Kaminski2018TheExplained} to provide an explanation of how data lead an automated algorithm, for example, to reject a loan application \cite{Kaminski2018TheExplained, kaminski2020algorithmic, Hacker2020ExplainableChallenges}. Finally, opaque models can conceal dataset bias, and lead to socially unfair models \cite{Shin2021TheAI}. 

In this work, we are particularly interested in explaining models in supervised learning scenarios in order to gain further insights about the concept related to a specific response. For example, assume one has a model that predicts the presence of brain tumor in MRI scans with very high accuracy. What are the most relevant morphological features that indicate the presence of tumor, and where are they located? Can we discover new features of the disease from what the model has learned? Many important problems of this kind exist, but the necessary tools to answer these questions effectively and efficiently are still lacking.

The foundational work by Ribeiro et al. \cite{Ribeiro2016WhyYou} spurred exciting advances in local feature attribution methods, such as Grad-CAM \cite{Selvaraju2016Grad-CAM:Localization}, Integrated Gradients \cite{Sundararajan2017AxiomaticNetworks}, and DeepLIFT \cite{Shrikumar2017LearningDifferences}. Lundberg and Lee \cite{Lundberg2017APredictions} provide a unified framework for several different approaches under their SHAP method, which leverages Shapley coefficients--a game-theoretic measure \cite{Shapley1953}--and feature removal strategies. Unlike other perturbation-based alternatives \cite{shah2021input}, these methods enjoy of important consistency results and theoretical properties that the resulting attributions satisfy. Since then, a plethora of different explanation methods has been developed\footnote{To our knowledge, Covert et al. \cite{covert2021explaining} compiled the most comprehensive review of currently available explanation methods based on feature removal.} for different kinds of data (tabular, sequential, imaging), both based on Shapley coefficients \cite{chen2018lshapley} as well as other information theoretic quantities \cite{MacDonald2019ADecisions, Hei2020In-DistributionModalities, Merrick2020Explanation}. Although previous work explores structured and hierarchical approaches \cite{Chen2020GeneratingDetection, chen2018lshapley, Singh2018HierarchicalPredictions}, they remain limited for high-dimensional data.

Notwithstanding the recent advances in image attribution methods based on Shapley coefficients, several limitations hinder their use for ``large'' images--a standard image contains $\approx 10^6$ pixels, and larger images are used in several important applications. We focus on problems that satisfy a certain \emph{multiple instance learning} assumption \cite{dietterich1997solving}, which can be found in many relevant fields. We show that in these problems, the computation of Shapley coefficients can be solved efficiently and without the need of approximation by exploring a hierarchical partition of the input image. The contribution of this work is three-fold: first, we present a fast explanation method based on Shapley coefficients that is exponentially faster than popular SHAP methods. Second, under some distributional assumptions similar to those in multiple instance learning problems, we show that the coefficients provided by h-Shap are exact, and can be further approximated in a controlled manner by trading off computational cost. Third, we compare h-Shap with other popular explanation methods on three benchmarks, of varied complexity and dimension, demonstrating that h-Shap outperforms the state of the art both in terms of runtime and retrieval of relevant features in all experiments.

This paper is organized as follows. In \cref{sec:brackground} we briefly summarize the necessary background. We present h-Shap in \cref{sec:H-SHAP}, including results on computational complexity and approximation. We present experiments in \cref{sec:experiments} and their results in \cref{sec:results}. Finally, we discuss our limitations in \cref{sec:limitations}, and we conclude in \cref{sec:discussion}.

\section{\label{sec:brackground}Background}

In supervised learning scenarios, we are interested in approximating a response or label, $Y \in \Y$, from a given input random sample $X \in \X$. Herein we assume a realizable setting where the response $Y = f^*(X) \in \Y$, for some $f^*: \X \to \Y$, and denote the joint distribution of $(X,Y)$ as $\mathcal{D}$. We look for a function $f:\mathcal X \to \mathcal Y'$ that approximates $f^*(X)$. Given a loss function $L: \Y \times \Y' \to \mathbb{R}$ that penalizes the dissimilarity between the predicted and real label, we look for $f$ in a suitable functional class with minimal risk, $\mathcal{R} = \mathbb{E}_{\mathcal{D}}[L(Y,f(X))]$. However, $\mathcal{D}$ is typically unknown and instead we are provided with a training set $\{(X^{(i)}, Y^{(i)})\}_{i=1}^N$ of observed data. As a result, we search for a function that minimizes the empirical risk,
\begin{equation}
    \f = \argmin_{f\in\mathcal{F}} \frac{1}{N}\sum_{i=i}^N L(Y^{(i)}, f(X^{(i)})),
\end{equation}
where $\mathcal F = \{f_\theta : \theta \in \Theta\}$, with parameters $\theta$ (such as a neural network model). We focus on binary classification problems, where  $\Y = \{0, 1\}$ and $\Y' \in \mathbb R$, though our general methodology is applicable to multi-class settings as well. We will refer to images--matrices of size $(\sqrt{n}\times\sqrt{n})$--as vectors in the $n$-dimensional real space, i.e. $\X \subseteq \mathbb{R}^n$. 

\subsection{Explaining predictions via Shapley coefficients}

Modern machine learning models, in particular those based on deep neural networks, can often provide solutions that perform remarkably well. In many settings, however, one would like to know the contribution of $x_i$, the $i^{th}$ entry of $X$, towards the output. Let us define by $C$ a subset of the entries of $X$, so that $C \subseteq [n] \coloneqq \{1,\dots,n\}$, and define $X_C \in\mathbb{R}^n$ the input that coincides with $X$ in the entries denoted by $C$ but takes a different, \emph{baseline}, value in its complement, $\bar{C}$. In the context of interpretability, we look for a vector $\Phi_{(X,\f)}\in\mathbb{R}^n$, where the $i^{th}$ coordinate reflects the importance of $x_i$ in producing the output $\hat{f}(X)$. Broadly speaking, the features in $C$ provide an explanation for the local prediction $\f(X)$ if $\f(X) \approx \f(X_C)$. Different measures of importance have been proposed to study model interpretability, and thus to compute $\Phi_{(X,\f)}$. In this work we focus on the general approach presented originally by \cite{Lundberg2017APredictions} that employs Shapley coefficients \cite{Shapley1953} as the measure of contribution of every pixel toward the output, which has gained great popularity \cite{Sundararajan2020Many}. We now briefly introduce some game theory notation to define Shapley coefficients.

Let $g = (X, f, [n])$ be an $n$-person cooperative game with players $[n]$ and characteristic function $f:\X \mapsto \mathbb{R}$ which maps the input space $\X$ to a score. In particular, $f(X_C)$ is the score that the players in $C$ would earn by collaborating in the game, with $f(X_\emptyset) = 0$ by convention\footnote{Formally speaking, game theory \cite{Owen1995GameTheory} requires a characteristic function $v: \mathcal{P}(X) \to \mathbb{R}$, where $\mathcal{P}(X)$ is the power set of $X$. Herein, and following prior work \cite{Lundberg2017APredictions}, we assume $v(C) = f(X_C),~\forall C \subseteq X$, and therefore use $f$ for the sake of simplicity.}. A \textit{solution concept} is a rule that assigns a fair contribution to each player in the game. Notably, Shapley coefficients, denoted by $\phi_1(f), \dots, \phi_n(f)$, are the only solution concept of $(X, f, [n])$ that simultaneously satisfy the properties of efficiency, linearity, symmetry, and nullity \cite{Shapley1953}. In the context of model explanations, input features are regarded as players, and these properties imply that: $\mathbf{i)}$ feature attributions sum up to the model prediction; $\mathbf{ii)}$ the attributions of features playing a convex combination of games are equal to the convex combination of the attributions of the features playing the individual games independently; $\mathbf{iii)}$ the attributions of irrelevant features are simply \num{0}; and $\mathbf{iv)}$ the attributions of equally important features are equal, respectively. These equip Shapley-based methods with a useful set of properties, which are not generally satisfied by others attributions methods \cite{shah2021input}. 

Shapley coefficients can be derived axiomatically \cite{Shapley1953}, and they are defined as
\begin{equation}
    \label{eq:definition}\phi_i(f) = \sum_{C \subseteq [n] \setminus \{i\}} \frac{\lvert C \rvert!(n -\lvert C \rvert - 1)!}{n!}  \left[f(X_{C \cup \{i\}}) - f(X_C)\right].\\
\end{equation}
This way, $\phi_i(f)$ represents the averaged marginalized contribution of $x_i$ over all possible subsets of $[n]$. \cref{eq:definition} also illustrates what is arguably the most important limitation of Shapley coefficients: their computational cost is exponential in the dimension of the input features (as there are exponentially many distinct subsets $C$) requiring $2^n$ unique evaluations of $f$. This quickly becomes intractable in image classification problems when $f$ is a convolutional neural network and $n \approx 10^6$, or larger.
As a result, all state-of-the-art image explanation methods based on Shapley coefficients rely on some approximation strategy to work around this computational limitation. For instance, GradientExplainer \cite{Lundberg2017APredictions} extends Integrated Gradients \cite{Sundararajan2017AxiomaticNetworks} by sampling multiple references from the background dataset to integrate on. Similarly, DeepExplainer \cite{Lundberg2017APredictions, Chen2021ExplainingComponents} builds upon DeepLIFT \cite{Shrikumar2017LearningDifferences} by choosing a per-node attribution rule that can approximate Shapley coefficients when integrated over many background samples. Finally, PartitionExplainer employs a hierarchical clustering approach inspired by Owen coefficients \cite{Owen1977ValuesUnions, Lopez2009OnValues, huettner2012axiomatic}, which generalize Shapley coefficients to cooperative games with \textit{a-priori} coalition structures. Given a game $(X, f, [n])$, let $\G = \{G_1, \dots, G_m\}$ be a coalition structure such that $\bigcup_{G \in \G} = [n]$, and $G_q \cap G_u = \emptyset$ for $q \neq u$. Then, the Owen coefficient of $x_i$ is defined as
\begin{equation}
    \label{eq:owen_definition}\varphi_i(f) = \sum_{\substack{H \subseteq [m] \setminus \{q^*\} \\ C \subseteq G_{q^*} \setminus \{i\}}} w_{H, M, C, G_{q^*}} \left[ f(X_{Q \cup C \cup \{i\}}) - f(X_{Q \cup C}) \right],
\end{equation}
where $w_{H, M, C, G_{q^*}}$ is an appropriate constant, $[m] \coloneqq \{1, \dots, m\}$, $Q = \bigcup_{q \in H} G_q$, and $i \in G_,~q^* \in [m]$. Similarly to Shapley coefficients, Owen coefficients are the only solution concept that satisfy similar properties of efficiency, marginality, and symmetry both across and within coalitions \cite{khmelnitskaya2007owen}. Intuitively, when looking at feature $i$ from the perspective of Shapley coefficients (i.e. \cref{eq:definition}), one has to consider all possible subsets of the remaining players. On the other hand, when considering the Owen coefficient of feature $i$ in coalition $G_{q^*}$ (i.e. \cref{eq:owen_definition}), one can only observe other coalitions participate in the game together as a whole, while still being able to observe all possible subsets of players within coalition $G_{q^*}$. This \emph{a-priori} coalition structure reduces the number of subsets of players to explore. Given the close relation between PartitionExplainer and h-Shap, we include a detailed comparison in \cref{supp:partitionexplainer_comparison}. 

To conclude, while the methods above provide approximations that can sometimes work in practice, they only provide consistency results and lack accuracy guarantees when they are run with a few model evaluations \cite{Merrick2020Explanation}. Hence, it is hard to understand when they will and will not be effective. We will compare extensively with these approaches later in \cref{sec:experiments}.

We remark that one of the most important details of any explanation method based on feature removal is the baseline, which defines the value that $X_C$ takes in the entries not in $C$. There are different approaches to removing features, ranging from using the default value of \num{0}, to using their conditional distribution (refer to \cite{covert2021explaining} for further details). Computing the latter can be challenging, and recent work has explored various approximations \cite{Aas2019ExplainingValues, Frye2019AsymmetricExplainability}. The effects of using different baselines have also been investigated in images \cite{Sturmfels2020VisualizingBaselines} and tabular data \cite{Haug2021OnAttributions}. We follow the standard approach of setting the baseline to the unconditional expected value over the training dataset \cite{Lundberg2017APredictions, Janzing2019FeatureProblem}, and comment on potential extensions later.

\subsection{Multiple Instance Learning}

In this work, we focus on problems with particular joint distributions of samples and labels. Our guarantees will apply to settings broadly known as \emph{Multiple Instance Learning} (MIL) \cite{dietterich1997solving, Weidmann2003A2003}. In MIL, each \emph{instance} $x_i$ is assumed to have an instance-label, and the sample $X$ is regarded as a \emph{bag} that aggregates all instances. The bag, $X$, has its own label $Y \in \{0,1\}$ determined by its constituent instances. In its simplest version, the bag is assumed to be positive if at least one of its instances is positive. As an example, an image of cells will be labeled with \texttt{infection} if at least one cell in it is \texttt{infected}. Importantly, the learner does not have access to the instance-labels, but only to the global label $Y$. Such an MIL setting appears in several important problems \cite{Han2020AccurateLearning, Hashimoto2020MultiscaleDomain, fu2012implementation}. In the context of our work, we assume that the prediction rule satisfies such an MIL assumption. More precisely, we will assume that
\begin{equation}
    \label{eq:MIL_assumption}
    f^*(X) = 1 \iff \exists~C \subseteq [n]:~f^*(X_C) = 1.
\end{equation}
In words, \cref{eq:MIL_assumption} implies that $f^*(X)$ will be \num{1} as soon as there is at least one subset $C$ of $[n]$ that contains the \emph{concept} we are interested in detecting. This is simply a formalization of the setting we were describing earlier, where the concept can be a specific morphological feature in a brain scan, a sick cell in a blood smear, or something as general as a traffic light in a street image.

As a partial summary of this section, $\f$ is trained to detect a binary concept in a sample image, and we would like to detect which subsets of the input, $X_C$, are relevant for this task. While this could in principle be done via Shapley coefficients, this is computationally intractable. We now move on to present our approach, which will address this limitation.

\section{\label{sec:H-SHAP}Hierarchical-Shap}

Our motivating observation is that if an area of an image is uninformative (i.e. it does not contain the concept), so will be its constituent sub-areas. Therefore, the exploration of relevant areas of an image can be done in a hierarchical manner. There exists extensive literature on hierarchies of games and their properties \cite{Faigle2008ATheory, Encarnacion2019TheHierarchies}. Our contribution is to deploy these ideas for the purpose of image explanations. 

We now make this more precise. Let $\T_0 = (S_0, \T_1, \dots, \T_{\gamma})$ be a recursive $\gamma$-partition tree of $X$, where $S_0$ is the root node containing all features of $X$, i.e. $S_0 = [n], \lvert S_0 \rvert = n$, and $\T_1, \dots, \T_{\gamma}$ are the subtrees branching off of $S_0$. Let $c(S_i) = \{C_1, \dots, C_\gamma\}$ denote the children of $S_i$, and $h_{\f}: S_i \mapsto (X, \f, c(S_i))$ be a mapping from the node $S_i$ of $\T_i$ to the $\gamma$-person cooperative game $(X, \f, c(S_i))$. Succinctly, $\mathcal{G}_0 = h_{\f}(\T_0)$ is a hierarchy of $\gamma$-person games, and we denote by $\phi_{i,1}(\f), \dots, \phi_{i,\gamma}(\f)$ the Shapley coefficients of $g_i \in \mathcal{G}_0$. In simpler words, we partition an image $X$ into \emph{a few disjoint components}, compute the Shapley coefficients $\phi_i$ of each component, and then partition further in a hierarchical manner. In particular, the number of such partitions per level (specified by $\gamma$) is very small: if $X$ is a one dimensional vector, we set $\gamma=2$ and $\T_0$ is a binary tree; when $X$ is a $(\sqrt{n} \times \sqrt{n})$ image, $\gamma=4$ and $\T_0$ is a quadtree. As a result, computing all $2^\gamma$ unique evaluations of $\f$ required for each game $(X, \f, c(S_i))$ is trivial. For images, each coefficient requires only \num{16} model evaluations. In fact, the remaining coefficients (for the same node) involve the same terms but in different permutations, so no extra model evaluations are needed. We have chosen to employ symmetric disjoint partitions in this work (i.e. halves for vectors, quadrants for images, etc) for simplicity only. More sophisticated (and potentially data-dependent) hierarchical partitions are possible as well. We will comment on this in the discussion.

Given such nested partitions, h-Shap relies on evaluating the resulting hierarchy of games while only visiting nodes that are relevant. More precisely, beginning at $S_0$, it computes the coefficients $\phi_{0,1}, \dots, \phi_{0,\gamma}$ of $g_0$. Under \cref{eq:MIL_assumption}, if any $\phi_{0,i}=0$, all features in the corresponding subtrees will also be irrelevant. As a result, they can be ignored altogether, and we only proceed by exploring the $S_i$ for which $\phi_i > 0$. This process finishes when all relevant leaves have been visited. In practice, we introduce two parameters to add flexibility. We set a relevance tolerance, $\tau$, which determines the threshold to be used to declare a partition relevant, and therefore expand on its subtrees. We further introduce a minimal feature size, $s$, that serves as a condition for termination. These two parameters are naturally motivated by application and easy to set. For example, it might not be that useful for a domain expert to know the exact pixel-level explanation of a given input. Rather, it would be more informative to have a coarser aggregation of the features that inform the model prediction. Later in this section, we will precisely characterize how the minimal feature size $s$ affects the dissimilarity between h-Shap's attributions and the exact Shapley coefficients. On the other hand, model deviations and noise in the input may result in positive coefficients very close to \num{0}. Requiring $\phi_i > \tau > 0$ provides control over the sensitivity of the method. Finally, when $\tau = 0, s = 1$, h-Shap simply explores all relevant nodes in $\T_0$ as described above. 

\begin{algorithm}[t]
\caption{Depth-first h-Shap ($d$h-Shap)}\label{alg:depth-first}
\begin{algorithmic}[1]
    \Procedure{$d$\text{h-Shap}}{$X, \T_0, \f$}\\
    \textbf{inputs:} image $X$, threshold $\tau\geq0$, trained model $\f$
    \State $g_0 \gets (X, \f, c(S_0))$
    \State $\phi_{0,1,} \dots, \phi_{0,\gamma} \gets \texttt{shap}(g_0)$
    \ForAll{$\phi_i$}
        \If{$\phi_i > \tau$}
            \If{$\lvert S_i \rvert \leq s$}
                \State \textbf{return} $S_i$ 
            \Else
                \State \textbf{return} \texttt{$d$h-Shap}($X, \T_i, \f$)
            \EndIf
        \EndIf
    \EndFor
    \EndProcedure
    \State $L \gets \texttt{$d$h-Shap}(X, \T_0, \f)$
\end{algorithmic}
\end{algorithm}

Fixed $\tau$ and $s$, h-Shap explores $\T_0$ starting from $S_0$, and it visits all relevant nodes $S_i: \phi_i > \tau, \lvert S_i \rvert \geq s$. This tree exploration can be naturally done in a depth-first or breadth-first manner; \cref{alg:depth-first} presents $d$h-Shap (depth-first h-Shap). Please refer to \cref{alg:breadth-first} in \cref{supp:algorithms} for $b$h-Shap (breadth-first h-Shap). The only difference between the two algorithms is that the former defines $\tau$ as an absolute value (e.g. \num{0}), whereas the latter does so relative to the pooled Shapley coefficients of all nodes at the same depth (e.g. $50^{th}$ percentile). Both algorithms return the set of relevant leaves $L \subseteq [n]$ with coefficients greater than $\tau$, and the saliency map $\widehat{\Phi}_{(X, \f)}$ is finally computed as
\begin{equation}
    \label{eq:Mhat}
    \widehat{\phi}_i =
    \begin{cases}
        1/\lvert L \rvert & \text{if}~ i \in L,\\
        0 & \text{otherwise}.
    \end{cases}
\end{equation}
This choice will ensure that $\widehat{\Phi}_{(X, \f)}$ is consistent with the exact Shapley attributions $\Phi_{(X, \f)}$ under the MIL assumption, as we will formalize shortly. 

To mask features out (i.e. as baseline), h-Shap uses their expected value (or \textit{unconditional distribution} \cite{Janzing2019FeatureProblem}) for simplicity, as done by other works \cite{covert2021explaining}. As pointed out by \cite{covert2021explaining, Lundberg2017APredictions}, this is valid under the assumptions of model linearity and feature independence\footnote{We refer to \cite{Chen2020TrueData, Sundararajan2020Many, Merrick2020Explanation, Janzing2019FeatureProblem} for recent discussion on the use of \textit{observational} vs \textit{interventional} conditional distributions in the context of removal-based explanation methods.}. Yet, as we will argue later in \cref{sec:discussion}, the feature independence property holds approximately in the cases we are interested in this work, whereas our MIL assumptions are enough to provide specific guarantees without requiring linearity of the model. We will also show in \cref{sec:experiments} that these assumptions are sufficient for h-Shap to work well in practice. More generally, our contribution is independent of the particular method employed for sampling the baseline, and follow-up work can employ better approximations of both the observational and interventional conditional distributions in appropriate tasks \cite{Chen2020TrueData}.

\subsection{Computational analysis}

The benefit of h-Shap relies in decoupling the dimensionality of the sample $X$ (i.e. $n$), from the number of players in each game (i.e. $\gamma$). As we will explain in this section, this leads to an exponential computational advantage over the general expression in \cref{eq:definition} in explaining $\f$. In the analysis that follows, we do not include the computation of the baseline value--which we assume fixed, see discussion in \cref{sec:discussion}--and we refer the reader to the proofs of all the results in this section to the \cref{supp:proofs}.
Let us denote by $\hat\T_0$ the subtree of $\T_0$ explored by h-Shap (i.e. the one with the visited nodes only). We will also assume in this section that $n$ is a power of $\gamma$ for simplicity of the expressions\footnote{Note that it is trivial to accommodate cases where this is not true.}. We begin by making the following remark.

\begin{remark}[Computational cost]
    Given $X \in \mathbb{R}^n$, h-Shap requires at most $2^\gamma k \log_{\gamma}(n)$ model evaluations, where $k$ is the number of relevant leaves in $\hat\T_0$.
\end{remark}

This result follows directly by noting that the cost of splitting each node is always $2^\gamma$, and by realizing that each important leaf takes, at most, $\log_\gamma(n)$ nodes, which is exponentially better than the cost of \cref{eq:definition}. The reader should recall that the number of internal nodes of a full and complete $\gamma$-partition tree is $(n - 1)/(\gamma - 1)$. Then, the above result is relevant whenever $k \log_{\gamma}{n} < (n - 1)/(\gamma - 1)$. This implies that further benefit is obtained whenever $k = \mathcal{O}(n/\log_{\gamma}{n})$, which is only a mild requirement in the number of relevant features. 

Moreover, it is of interest to know the expected computational cost, which can be significantly smaller than the upper bound above. Throughout the rest of this section, and to provide more precise results, we will let the data $X$ be drawn from a distribution of \emph{important} and \emph{non-important} features. A distribution is ``important'' in the sense that it leads to positive responses.

\begin{assumption}
    \label{assump:A1}
    The data $X \in \mathbb{R}^n$ is drawn so that each entry $x_i \sim a_i \mathcal{I}~+~(1 - a_i)\mathcal{I}^c$, where $a_i \sim \text{Bernoulli}(\rho)$ is a binary random variable that indicates whether the feature $x_i$ comes from an \emph{important} distribution $\mathcal{I}$, or its \emph{non-important} complement $\mathcal{I}^c$, so that
    \begin{equation}
        \label{eq:assumption_mil_equation}
        \f(X_C) = 1 \iff \exists i \in C:~x_i \sim \mathcal{I},~C \subseteq [n].
    \end{equation}
\end{assumption}

With these elements, we present the following result.

\begin{figure}[t]
    \centering
    \includegraphics[width=0.5\linewidth]{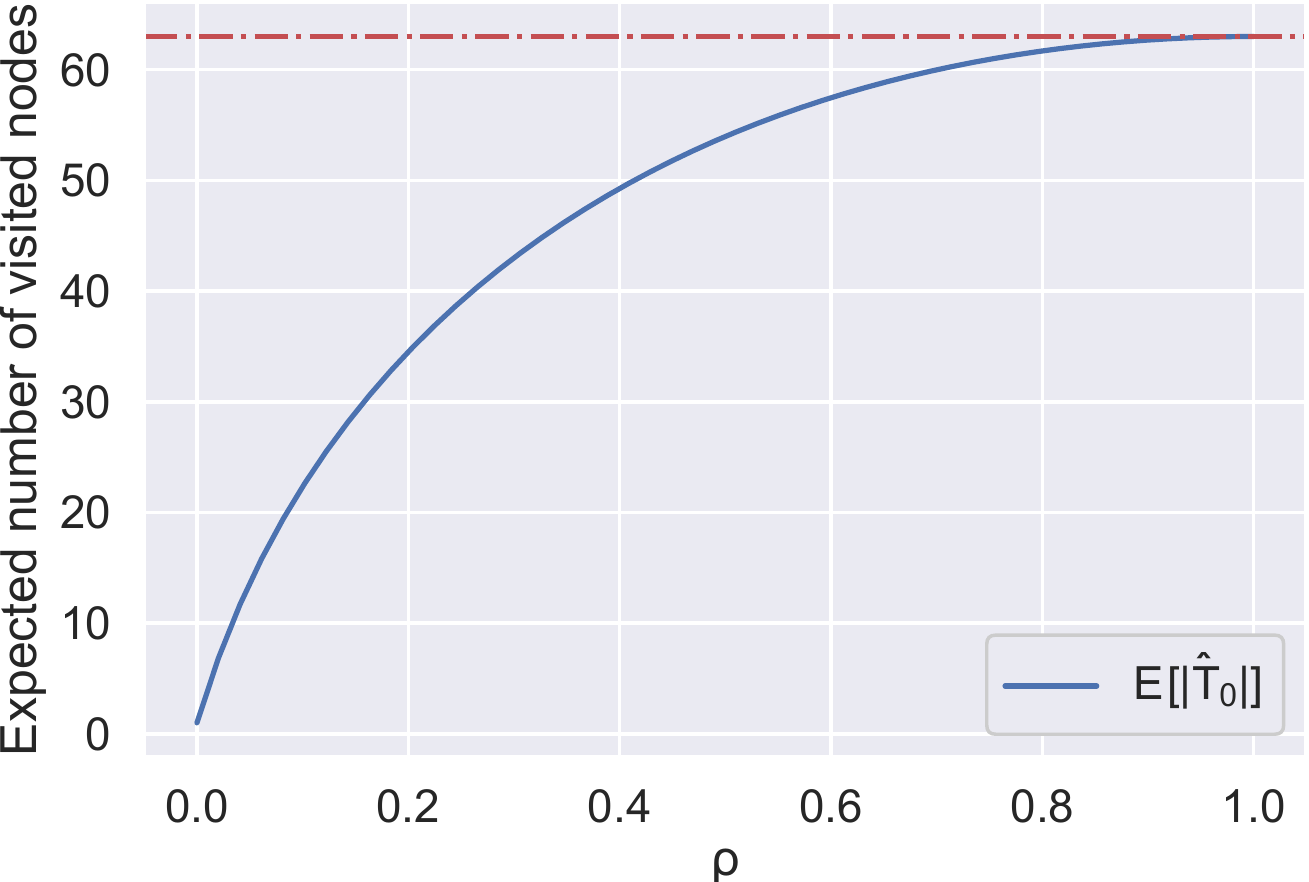}
    \caption{\label{fig:expected_visited_nodes}Expected number of visited nodes as a function of $\rho$ when $n=64, \gamma=2, s=1$.}
    \vspace{-10pt}
\end{figure}

\begin{theorem}[Expected number of visited nodes]
    \label{th:expectation}
    Assume $X$ and $\f(X)$ satisfy \cref{assump:A1}, $\tau = 0$, and $s = 1$. Then, the expected number of visited nodes in $\hat\T_0$ is
    \begin{equation}
        \label{eq:probability_split}
        \mathbb{E}[\lvert \hat\T_0 \rvert] = 1 + \gamma(1 - p(S_0)) \mathbb{E} [\lvert \hat\T_1 \rvert],
    \end{equation}
    where
    \[
        p(S_i) =
        \begin{cases}
            (1 - \rho)^{\frac{\lvert S_i \rvert}{\gamma}} & \text{if}~ i = 0,\\
            (1 - \rho)^{\frac{\lvert S_i \rvert}{\gamma}}\left(\frac{1 - (1 - \rho)^{\lvert S_i \rvert \frac{\gamma - 1}{\gamma}}}{1 - (1 - \rho)^{\lvert S_i \rvert}}\right) & \text{otherwise}.
        \end{cases}
    \]
\end{theorem}
See \cref{proof:expectation}. This result does not provide a closed-form expression for the expected number of visited nodes (and, correspondingly, computational cost), but it does provide a simple recurrent formula that can be easily computed. Naturally, this cost depends on the Bernoulli probability $\rho$, the average number of important features in $X$. We present the resulting $\mathbb E[\lvert\hat\T_0\rvert]$ for a specific case in \cref{fig:expected_visited_nodes} as a function of $\rho$, showing that indeed the expected cost can be much lower than the worst-case bound. While this result (and, centrally, \cref{assump:A1}) was presented for the case where the relevant features are of size \num{1}, similar results can be provided for the case when the minimal features size $s > 1$.

\begin{figure*}
    \centering
    \subcaptionbox{\label{fig:demos_synthetic}Synthetic dataset.}
    {\includegraphics[width=\textwidth]{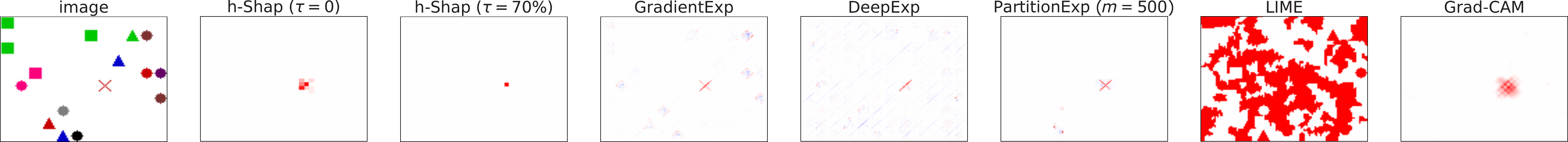}}
    \hfill
    \subcaptionbox{\label{fig:demos_malaria}BBBC041 dataset.}
    {\includegraphics[width=\textwidth]{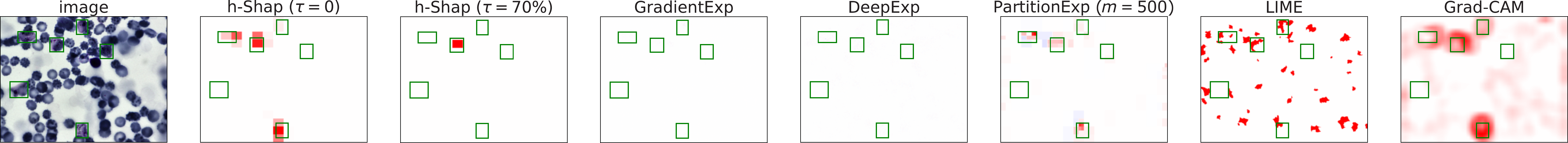}}
    \hfill
    \subcaptionbox{\label{fig:demos_lisa}LISA dataset.}
    {\includegraphics[width=\textwidth]{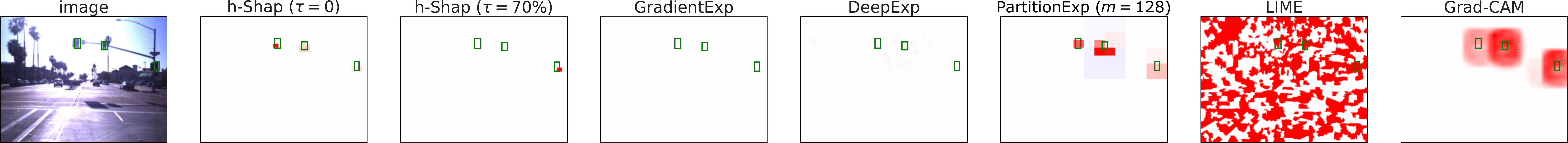}}
    \caption{\label{fig:demos}A few saliency maps for the three settings studied in this work, where blue pixels have negative, white pixels have negligible, and red pixels have positive Shapley coefficients. The color mapping is adapted to each saliency map and centered around \num{0}. For h-Shap, we show the saliency map before the normalization step.}
\end{figure*}

\subsection{Accuracy and Approximation}

Recall that h-Shap provides image attributions by means of a hierarchy of collaborative games. As a result, the attributions are different, in general, from those estimated by analyzing the grand coalition directly--that is, by the general Shapley approach in \cref{eq:definition}. We remark that computing the Shapley coefficients directly from \cref{eq:definition} quickly becomes intractable in image classification tasks. For example, even for a toy-like dataset of small $10 \times 10$ pixels images, assuming that each model computation takes $1$ nanosecond (which is unrealistically fast), computing the exact Shapley coefficients would take $\approx 3 \times 10^{13}$ years. Yet, we now show that under \cref{assump:A1}, h-Shap can in fact provide exact Shapley coefficients while being exponentially faster. Additionally, h-Shap can provide controlled approximations by trading computational efficiency with accuracy. 

We begin by noting that under the MIL assumption, all positive features have the same importance. This agrees with intuition that the number of times the positive concept appears in the input image does not affect its label. We denote as $\Phi$ and $\widehat\Phi$ the exact and hierarchical Shapley coefficients, respectively, for simplicity. 
\begin{remark}
    Under \cref{assump:A1}, and denoting $k=\|\Phi\|_0$, it holds that the exact saliency map $\Phi$ satisfies
    \begin{equation}
        \label{eq:exact_shap_saliency}
        \phi_i = 
        \begin{cases}
            1 / k & \text{if}~ x_i \sim \mathcal{I}\\
            0 & \text{otherwise}.
        \end{cases}
    \end{equation}
\end{remark}
This remark follows simply from the nullity and symmetry properties of Shapley coefficients. As a result, the saliency map computed by h-Shap, $\widehat\Phi$, as in \cref{eq:Mhat}, coincides with $\Phi$ under the MIL assumption. We now derive a more general similarity lower bound between $\Phi$ and $\widehat\Phi$ that allows for minimal feature sizes $s > 1$. For simplicity, we assume that $n$ and $s$ are powers of $\gamma$, and $1 \leq s \leq n$. First of all, because of the MIL assumption, h-Shap will \emph{only} keep exploring nodes that have at least one important feature in them at each level of the hierarchy. Thus, for each important feature $i$ with $\Phi_i = 1/k$ there will be a non-zero coefficient produced by h-Shap. The following result precisely quantifies to what extent these two vectors $\Phi$ and $\widehat{\Phi}$ match.

\begin{theorem}[Similarity lower bound]
    \label{th:cosine_similarity}
    Assume $X \in\mathbb R^n$ and $\f(X)$ satisfy \cref{assump:A1}, and $k=\|\Phi\|_0$. Then
    \begin{equation}
        \frac{\langle \Phi , \widehat\Phi \rangle}{\|\Phi\|_2 \|\widehat\Phi\|_2} \geq \max\{1/\sqrt{s}, \sqrt{k/n}\}.
    \end{equation}
\end{theorem}

See \cref{proof:cosine_similarity}. This result shows that not only does h-Shap provide faster image attributions, but it retrieves the exact Shapley coefficients defined in \cref{eq:exact_shap_saliency} under the MIL assumption if $s=1$. Notwithstanding, one can employ a larger minimal feature size, $s>1$, while still providing attributions that are similar to the original ones. In light of the result in \cref{th:expectation}, the latter attributions will naturally result in improved (smaller) computational costs.

\section{\label{sec:experiments}Experiments}

\begin{figure*}
    \centering
    \subcaptionbox{\label{fig:ablations_original}}
    {\includegraphics[width=0.14\linewidth]{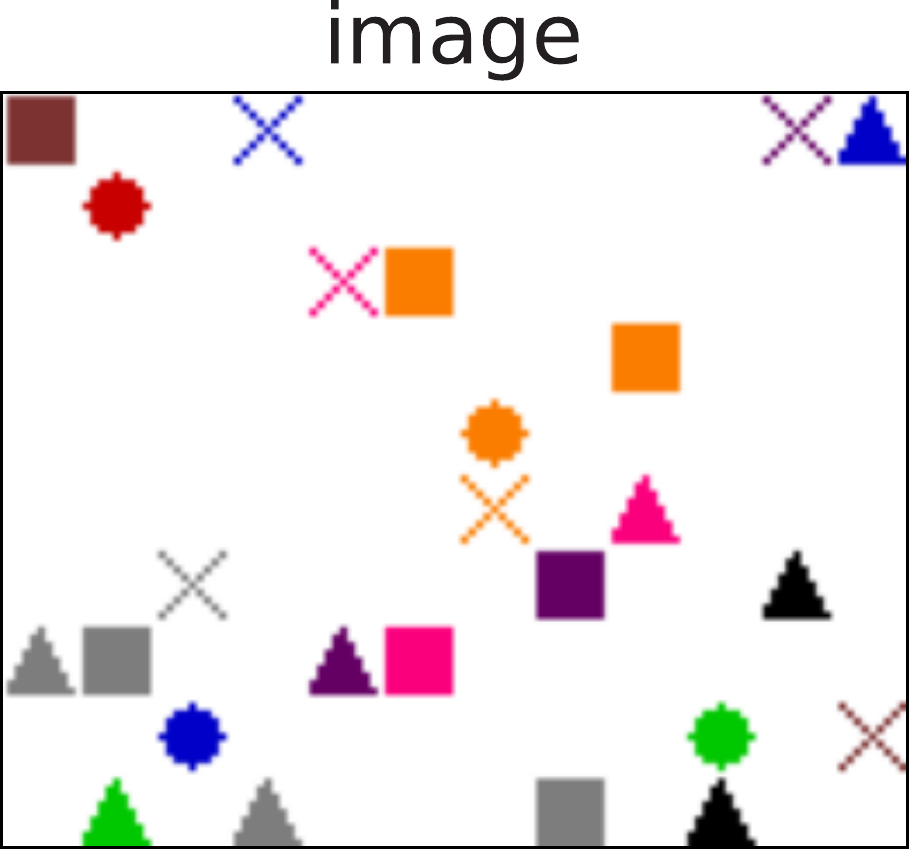}}\hfill
    \subcaptionbox{\label{fig:ablations_dhshap}}
    {\includegraphics[width=0.14\linewidth]{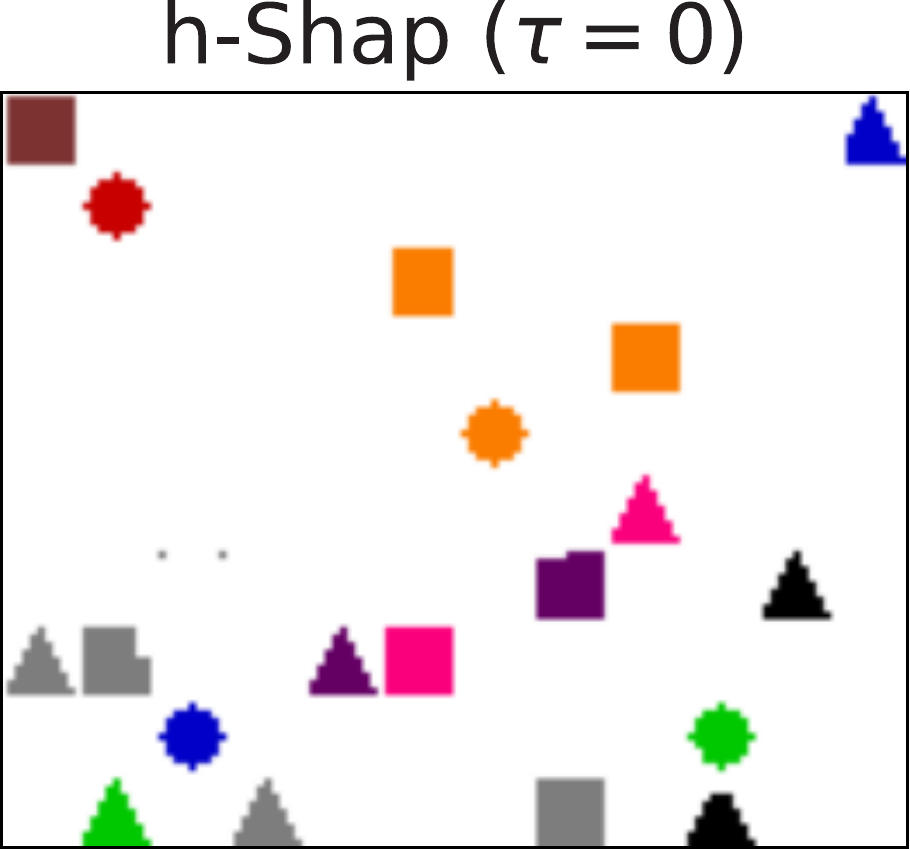}}\hfill
    \subcaptionbox{\label{fig:ablations_gradexp}}
    {\includegraphics[width=0.14\linewidth]{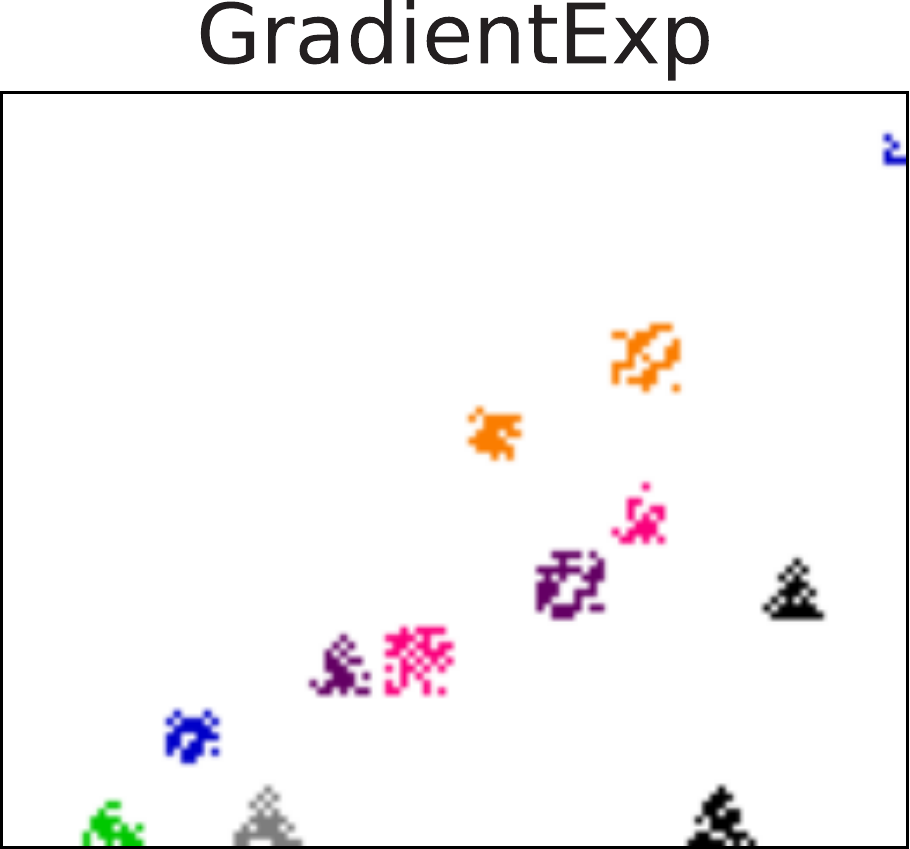}}\hfill
    \subcaptionbox{\label{fig:ablations_deepexp}}
    {\includegraphics[width=0.14\linewidth]{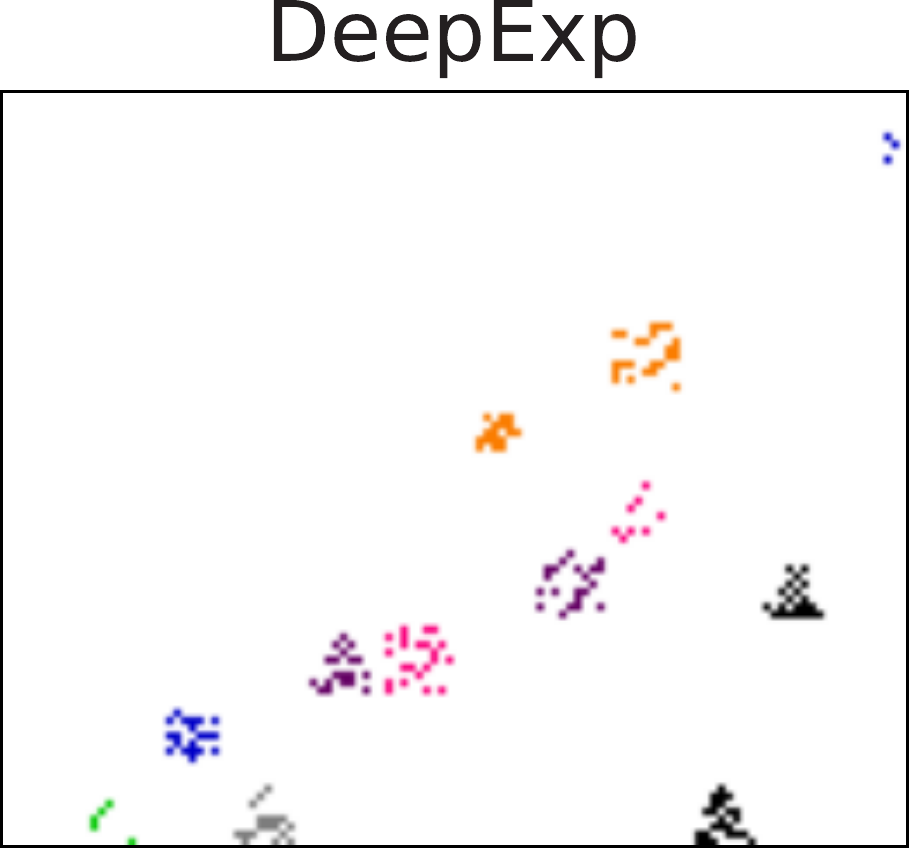}}\hfill
    \subcaptionbox{\label{fig:ablations_partexp}}
    {\includegraphics[width=0.14\linewidth]{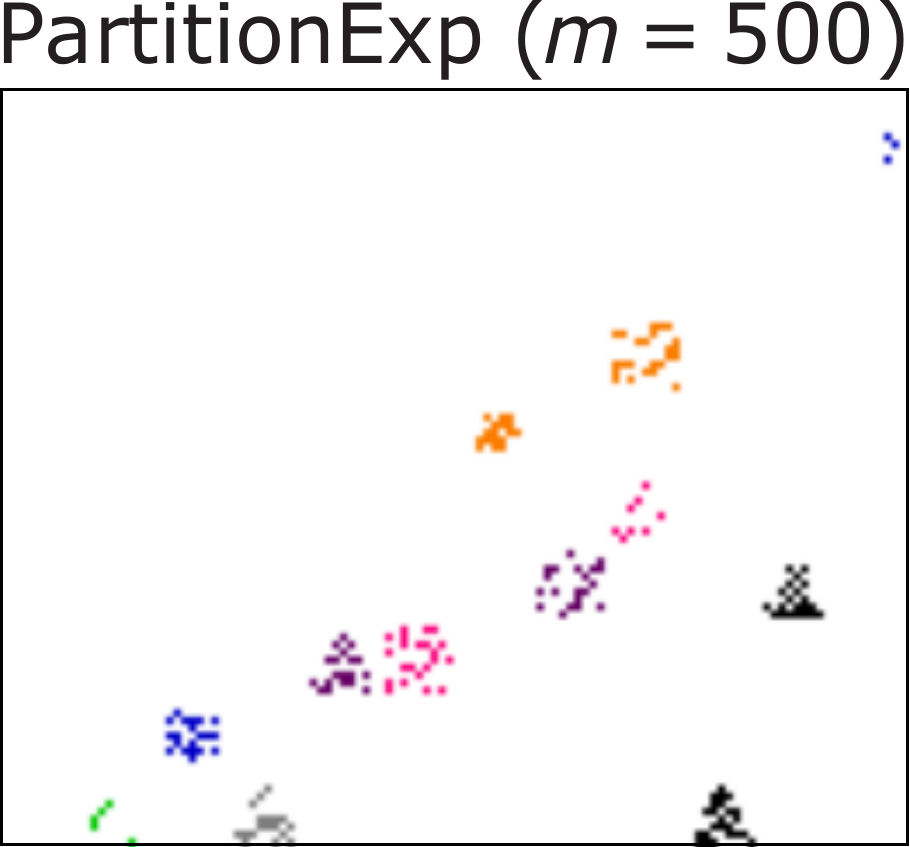}}\hfill
    \subcaptionbox{\label{fig:ablations_lime}}
    {\includegraphics[width=0.14\linewidth]{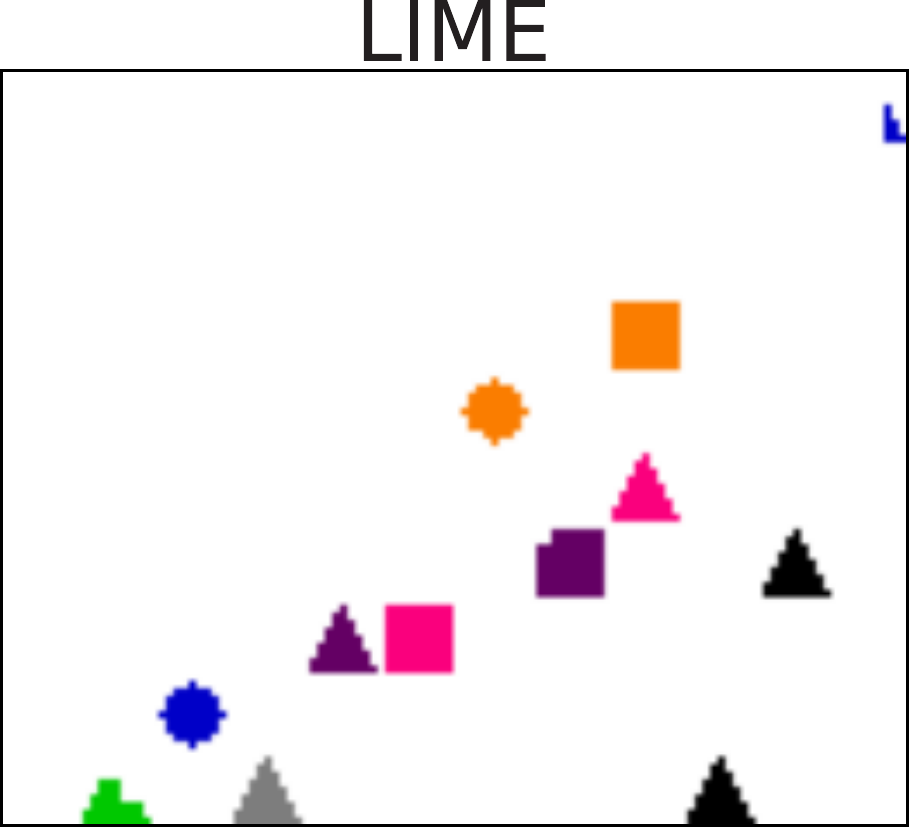}}\hfill
    \subcaptionbox{\label{fig:ablations_gradcam}}
    {\includegraphics[width=0.14\linewidth]{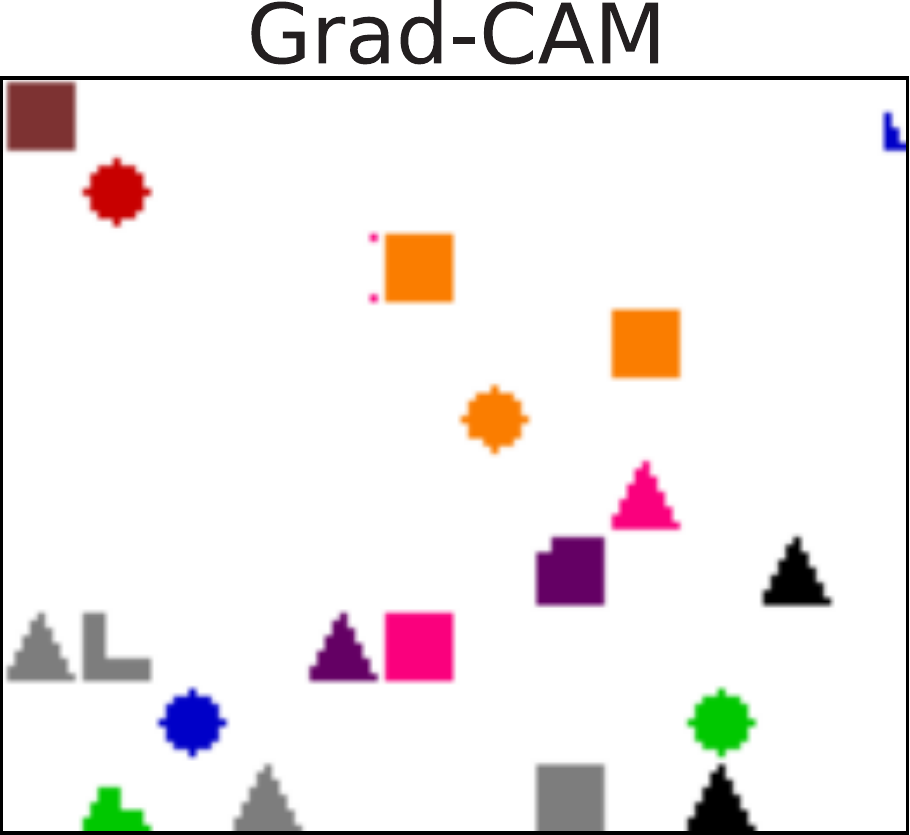}}
    \captionsetup{subrefformat=parens}
    \caption{\label{fig:ablations}Ablation examples for all explanation methods removing all important pixels from the original image \subref{fig:ablations_original}. The model is trained to predict if a given image does contain a cross or not.}
\end{figure*}

We now move to demonstrate the performance of h-Shap and of other state-of-the-art methods for image attributions. Our objective is mainly to compare with other Shapley-based methods, such as GradientExplainer \cite{Lundberg2017APredictions}, DeepExplainer \cite{Lundberg2017APredictions, Chen2021ExplainingComponents}, and PartitionExplainer\footnote{\label{footnote:shap}The implementation of GradientExplainer, DeepExplainer and PartitionExplainer are openly available at {https://github.com/slundberg/shap.}}. We also include LIME\footnote{{https://github.com/marcotcr/lime}.} \cite{Ribeiro2016WhyYou} given its relation to Shapley coefficients, and Grad-CAM\footnote{{https://github.com/jacobgil/pytorch-grad-cam}.} \cite{Selvaraju2016Grad-CAM:Localization} because of its popularity. We study three complementary binary classification problems of different complexity and input dimension: a simple synthetic benchmark, a medical imaging dataset, and a general computer vision task. We focus on scenarios where the ground truth of the image attributions (i.e. what defines the label) is well defined and available for evaluation. All experiments were conducted on a workstation with NVIDIA Quadro RTX 5000. Our code is made available for the purpose of reproducibility\footnote{https://github.com/Sulam-Group/h-shap}. When possible, each method was set to use as much GPU memory as possible, so as to minimize their runtime. DeepExplainer and GradientExplainer were constrained the most by memory, reflecting their limitation in analyzing large images. We use h-Shap with both an absolute threshold $\tau = 0$, and a relative threshold $\tau$ equal to the $70^{th}$ percentile, which we refer to as $\tau = 70\%$ with abuse of notation. Finally, we perform \textit{full} model randomization sanity checks \cite{Adebayo2018SanityMaps} on the network used in the synthetic dataset for all explanation methods. We refer the reader to \cref{supp:sanity_checks} for these results.

\subsection{Synthetic dataset} 

We created a controlled setting where the joint data distribution is completely known, giving us maximal flexibility for sampling. We generate images of size $100 \times 120$ pixels with a random number of non-overlapping geometric shapes of size $10 \times 10$ and of different colors, uniformly distributed across the image. Each image that contains at least one cross receives a positive label, and each image without any crosses receives a negative label. Alongside with the images, we generate the ground truth saliency maps by setting all pixels that precisely lie on a cross to \num{1}, and every other pixel to \num{0}. We generate \num{8000} positive and negative images, and we randomly sample train, validation, and test splits, with size $5000, 1000$ and $2000$ images, respectively. We train a simple ConvNet architecture, optimizing for \num{50} epochs with Adam \cite{Kingma2014Adam:Optimization}, learning rate of $0.001$ and cross-entropy loss. We achieve an accuracy greater than \SI{99}{\percent} on the test set--implying that the model has effectively satisfied the MIL assumption for this problem. From the true positive predictions on the test set, we choose \num{300} example images with \num{1} cross and as many with \num{6} crosses to evaluate the saliency maps. \cref{fig:demos_synthetic} presents a qualitative demonstration of h-Shap and other related methods on this task.

\subsection{P. vivax (malaria) dataset} 

Moving on to a real and high-dimensional problem, we explore the BBBC041v1 dataset, available from the Broad Bioimage Benchmark Collection\footnote{https://www.kaggle.com/kmader/malaria-bounding-boxes.} \cite{Ljosa2012AnnotatedValidation}. The dataset consists of \num{1328}, $1200 \times 1600$ pixels blood smears with uninfected (i.e. red blood cells and leukocytes) and infected (i.e. gametocytes, rings, trophozoites, and schizonts) blood cells. The dataset also comprises bounding-box annotations of both healthy and sick cells. We consider the binary problem of detecting images that contain at least one trophozoite, yielding \num{655} positive and \num{673} negative samples. Given the small amount of data available, we augment the training dataset with random horizontal flips, and we randomly choose \num{120} positive, and equally many negative images as the testing set. We apply transfer learning to a ResNet18 \cite{He2015DeepRecognition} network pretrained on ImageNet. We optimize all parameters of the pretrained network for \num{25} epochs with Adam \cite{Kingma2014Adam:Optimization}--learning rate \num{0.0001}. We use cross-entropy loss and learning rate decay of \num{0.2} every \num{10} epochs. After training, our model achieves a test accuracy greater than $\SI{99}{\percent}$. We finally aggregate all $112$ true positive predictions for evaluation, without distinction on the number of trophozoites in the image. \cref{fig:demos_malaria} shows a sample image and the corresponding saliency maps produced by the various methods.

\subsection{LISA traffic light dataset.} 

We finally look at a general computer vision dataset consisting of driving sequences collected in San Diego, CA, available from\footnote{https://www.kaggle.com/mbornoe/lisa-traffic-light-dataset.} \cite{Jensen2016VisionPerspectives, Philipsen2015TrafficDataset}. The complete dataset counts \num{43007} frames of size $960 \times 1280$ pixels, and \num{113888} annotated traffic lights. From this set, we take daytime traffic images, and train a model to predict the presence of a green light in a sample image. We respect the original train/test splits, providing \num{6108} train, \num{3846} test positive samples, and \num{6667} train, \num{3627} test negative samples. As before, we use data augmentation and apply transfer learning on a pretrained ResNet18. We optimize all parameters of the pretrained network for \num{25} epochs with Adam \cite{Kingma2014Adam:Optimization}--learning rate \num{0.0001}. We use cross-entropy loss and learning rate decay of \num{0.2} every \num{10} epochs. After training, we achieve a test accuracy of $\approx\SI{95}{\percent}$. Finally, we randomly sample \num{300} true positive examples to evaluate the different attribution methods on. \cref{fig:demos_lisa} illustrates a positive sample image, and the corresponding saliency maps.

\begin{figure*}[t]
    \centering
    \begin{subfigure}{0.9\linewidth}
        \includegraphics[width=0.53\linewidth]{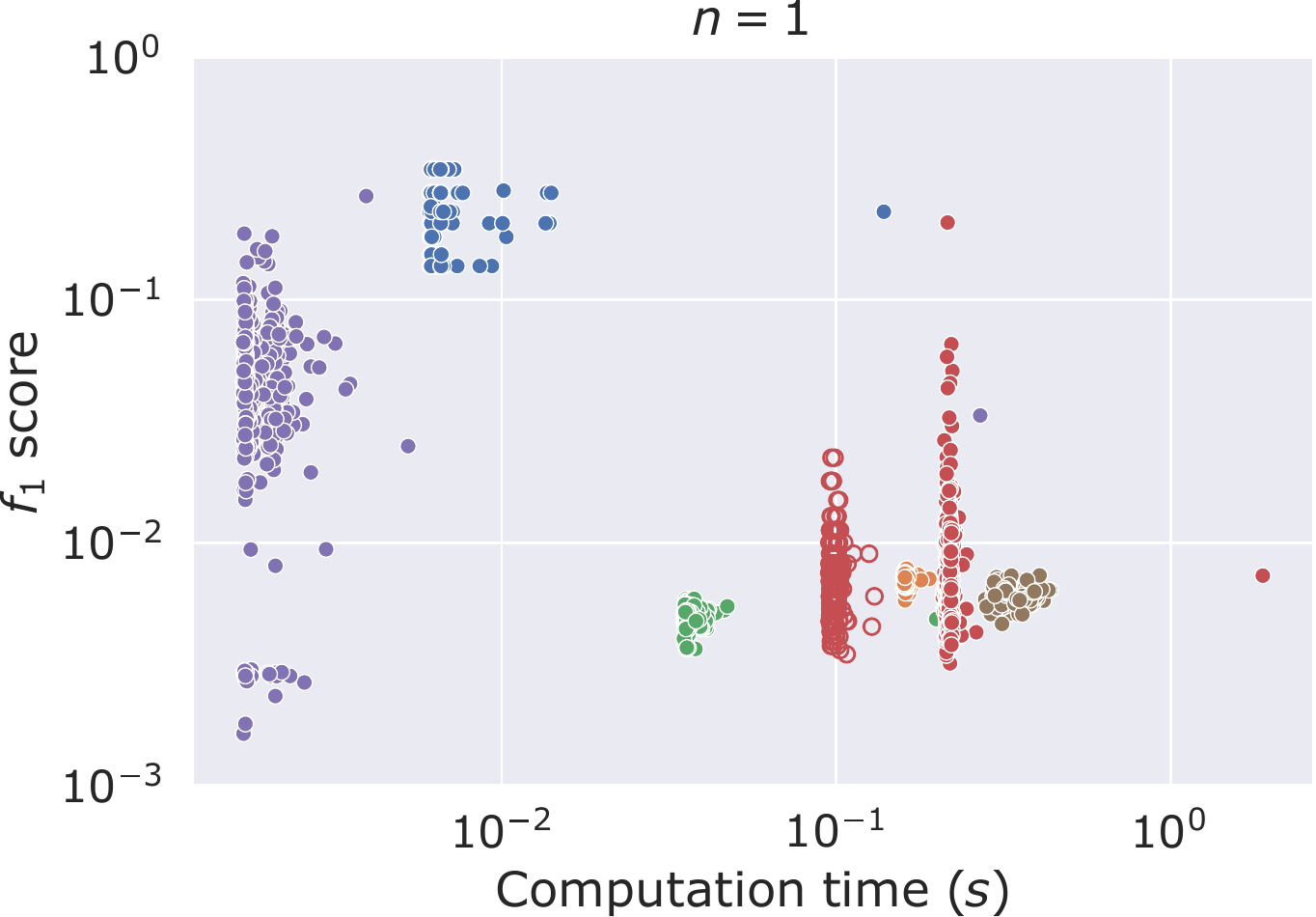}
        \includegraphics[width=0.455\linewidth]{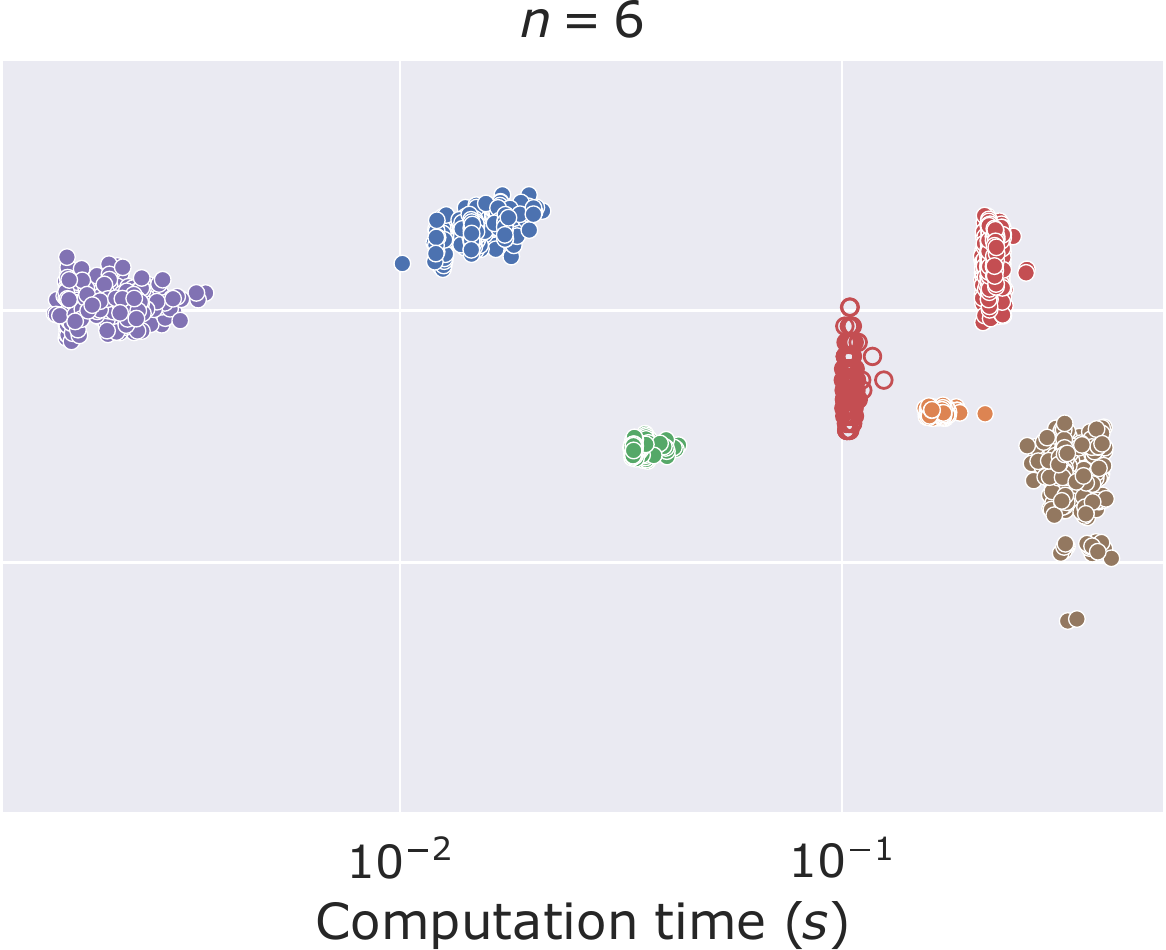}
        \caption{\label{fig:synthetic_results_f1}Synthetic dataset. Results for $n=1, 6$ crosses.}
    \end{subfigure}
    
    \begin{subfigure}{0.9\linewidth}
        \subcaptionbox{\label{fig:malaria_results_f1}BBBC041 dataset.}
        {\includegraphics[width=0.53\linewidth]{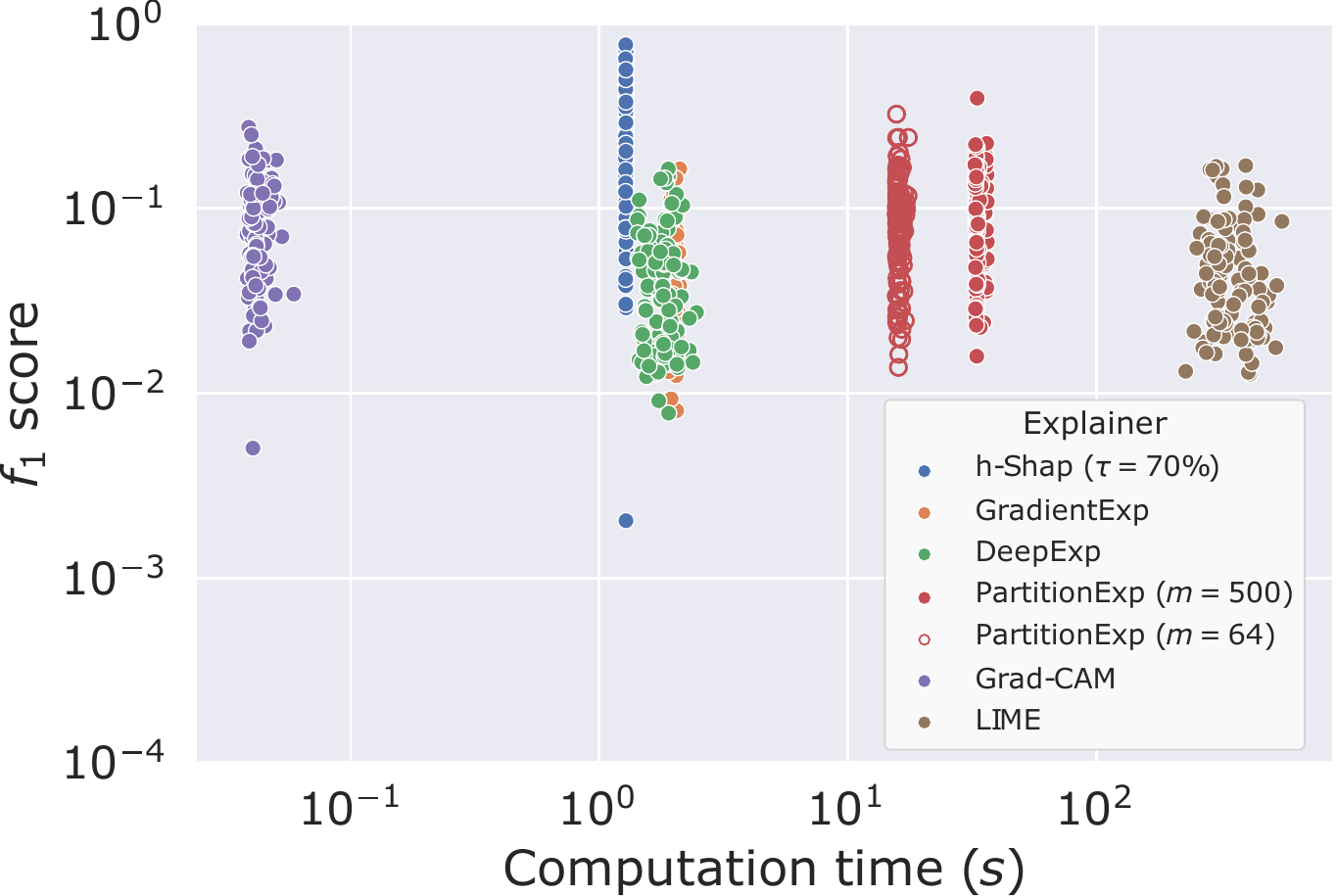}}
        \subcaptionbox{\label{fig:lisa_results_f1}LISA dataset.}
        {\includegraphics[width=0.455\linewidth]{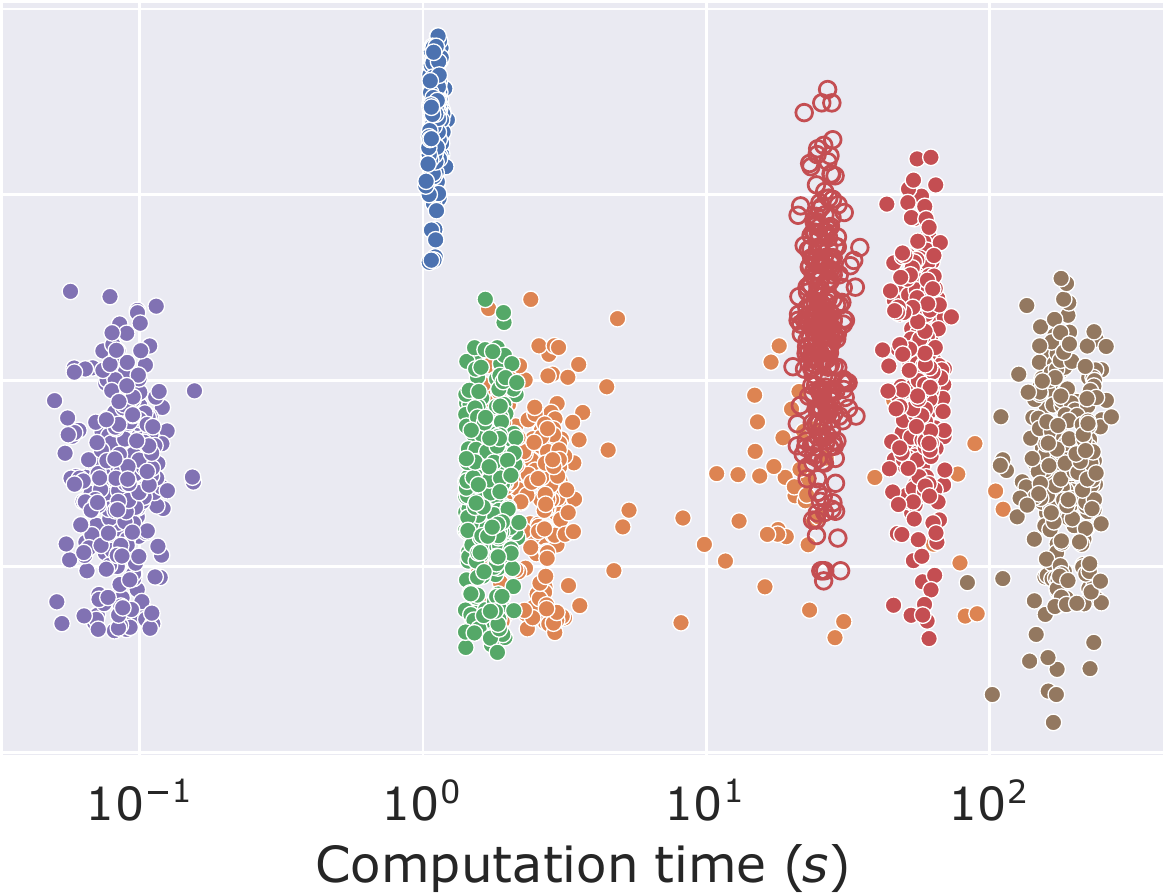}}
    \end{subfigure}
    \vspace{-5pt}
    \caption{\label{fig:results_f1}$f_1$ scores as a function of runtime for all explanation methods in all three experiments. To account for noise in the explanations, we threshold saliency maps at $1 \times 10^{-6}$ and compute $f_1$ scores on the resulting binary masks. For PartitionExplainer, $m$ indicates the maximal number of model evaluations.}
    \vspace{-10pt}
\end{figure*}

\section{\label{sec:results}Results}

\cref{fig:demos} shows a visual comparison of some saliency maps obtained in the three experiments (for more examples, see \cref{fig:more_demos}). Note that while the saliency maps produced by GradientExplainer and DeepExplainer appear empty in \cref{fig:demos_malaria,fig:demos_lisa}, they are not, and instead the single pixels are too small to be visible (these are large images). This illustrates how current Shapley-based explanation methods fall short of producing informative saliency maps in problems with large images. We further evaluate the explanation methods by means of three performance measures: ablation tests, accuracy, and runtime.

\subsection{Ablation tests} 

As commonly done in literature \cite{Lundberg2017APredictions, Sturmfels2020VisualizingBaselines, Haug2021OnAttributions} we remove the top $k$ scoring features of all methods by setting them to their expected value, and plot the logit of the prediction as a function of $k$. For these experiments, we use $\tau = 0$ so as to find \textit{all} the features that are relevant for the model. \cref{fig:ablations} shows ablation results on one example image from the synthetic dataset for all explanation methods. We expect a perfect method to remove all crosses from the image--and only those. We can appreciate how h-Shap removes mostly only the crosses, while other methods also erase other shapes which should not be identified as important. Furthermore, removing more relevant features should produce a steeper drop of the prediction logit. We include the respective curves in \cref{fig:results_ablation}, depicting that h-Shap's logit curves either quickly drop towards \num{0} or provide a logit $\approx 0$ at complete ablation. Indeed, h-Shap quickly identifies the most relevant features in the image. Naturally, as tasks become harder, the accuracy of $\f$ decreases, and the model gets further away from the oracle function $f^*$. In these cases (for the real datasets), $\f$ might not satisfy \cref{eq:MIL_assumption}, resulting in noisier saliency maps, and correspondingly, non-monotonic curves.

\subsection{Accuracy and Runtime} 

Since we have ground-truth explanations in all these cases (i.e. a cross, a sick cell, or a green traffic light), we use $f_1$ scores as a measure of goodness of explanation. We argue that $f_1$ scores are a particularly informative measure for explanations (when ground-truth is known), and consistent with previous work \cite{Guidotti2021EvaluatingTruth}. 
\cref{fig:results_f1} depicts the $f_1$ scores as a function of runtime for every explanation method and experiment. The advantage of setting a relative relevance tolerance $\tau$ is clear: to detect the \emph{most} relevant features and discard the noisy ones, taking into account the risk of the model $\f$, while also decreasing runtime. These results reflect how the computational cost and accuracy guarantees described earlier translate into application. Not only does h-Shap decrease runtime compared to current Shapley-based explanation methods--by one to two orders of magnitude--but it also increases the $f_1$ score. \cref{fig:synthetic_results_f1} shows that h-Shap's accuracy is not affected by the number of crosses in the image, while other methods deteriorate when there is only one cross to detect in the image. Importantly, in all experiments--both synthetic and real--h-Shap consistently provides more accurate and faster saliency maps compared to other Shapley-based methods, and it is only beaten in speed by Grad-CAM, which provides less accurate saliency maps.

\begin{figure*}
    \centering
    \subcaptionbox{\label{fig:degradation_original}Original image.}
    {\hspace{-10pt}\includegraphics[width=0.2\linewidth]{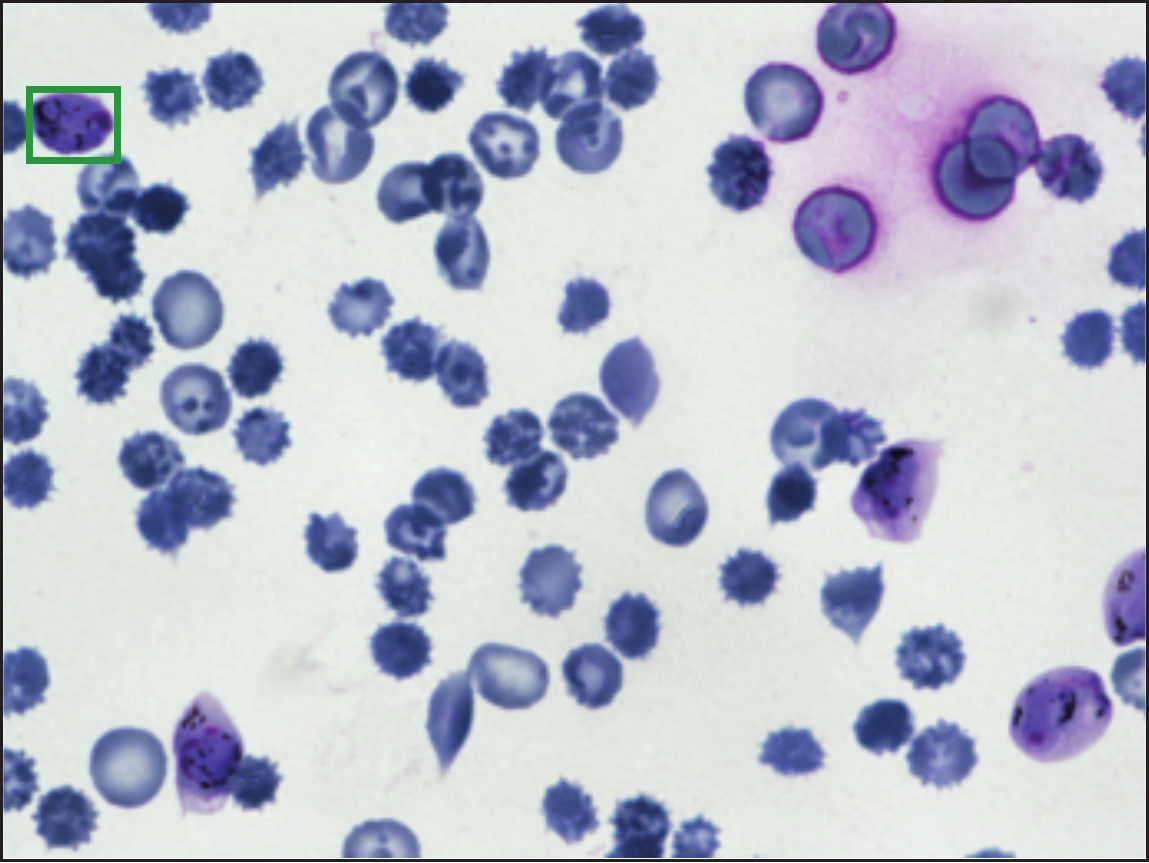}\hspace{60pt}}
    \subcaptionbox{\label{fig:degradation_40}$s = 80$ pixels.}
    {\includegraphics[width=0.2\linewidth]{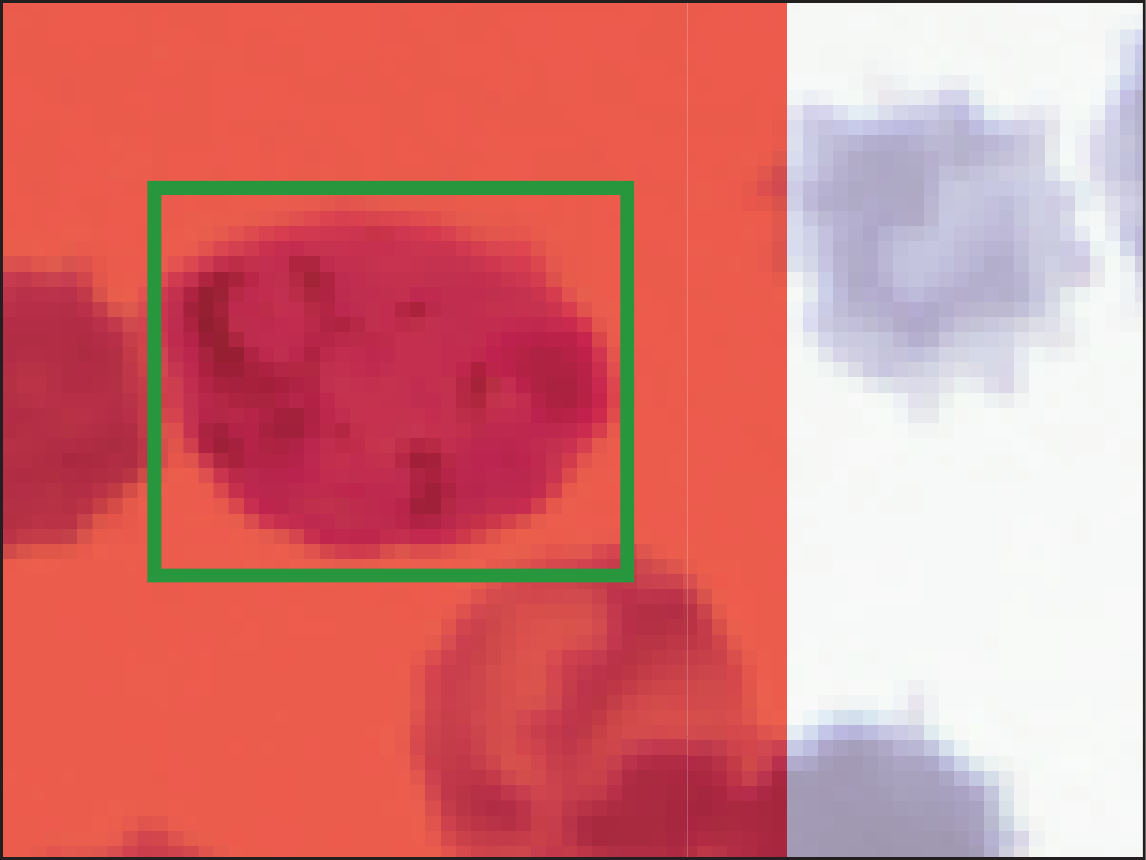}}
    \hspace{5pt}
    \subcaptionbox{\label{fig:degradation_20}$s = 20$ pixels.}
    {\includegraphics[width=0.2\linewidth]{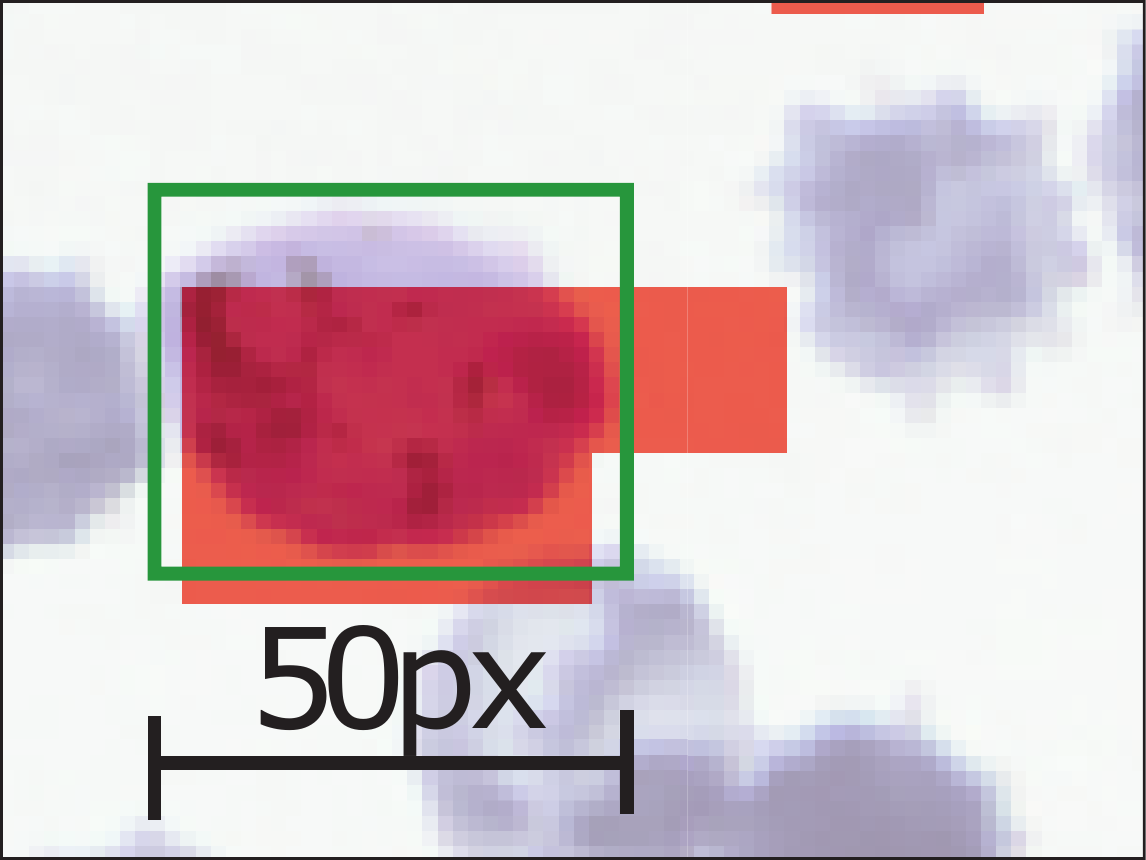}}
    \hspace{5pt}
    \subcaptionbox{\label{fig:degradation_5}$s = 5$ pixels.}
    {\includegraphics[width=0.2\linewidth]{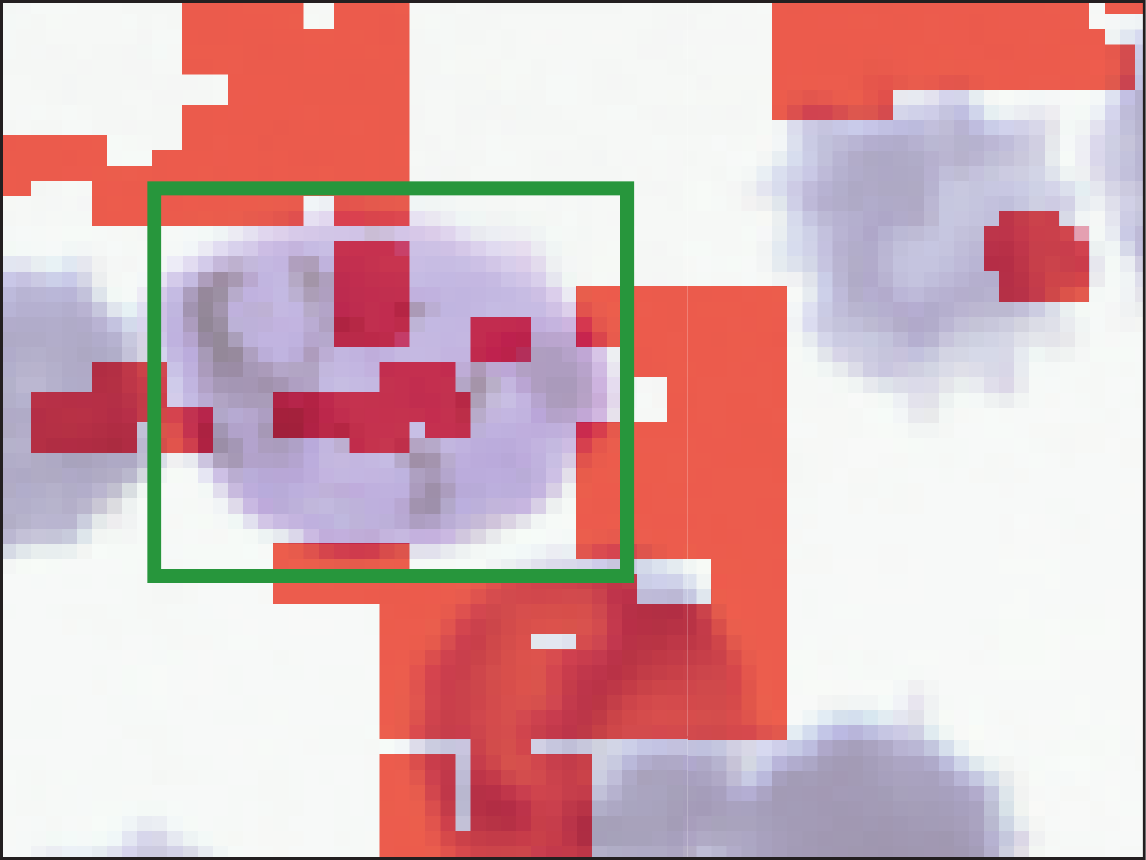}}
    \caption{\label{fig:degradation}Degradation of h-Shap's maps as the minimal feature size $s$ becomes smaller than the target concept.}
    \vspace{-15pt}
\end{figure*}

\section{\label{sec:limitations}Discussion}

\subsection{Limitations}
Before concluding, we want to delineate the limitations of h-Shap, the most important of which is its MIL assumption on the data distribution. The methodology proposed in this work is designed to identify local \emph{findings} that produce a positive global response, accurately and efficiently. These are precisely the important features $C$ analyzed in \cref{sec:H-SHAP}. This setting is controlled by the ratio of the size of the actual object that defines the label, and the minimal feature size of the algorithm. As an example, \cref{fig:degradation} depicts a zoomed-in version of the map produced by h-Shap for one of the samples from the P. vivax dataset, for different values of $s$. We see that even when $s$ is somewhat smaller than the object, h-Shap still recognizes the important features in the image. Once $s$ is too small, however, the resulting map breaks down, as our assumption does not hold any more. Indeed, small ($5 \times 5$ pixels) image patches break \cref{assump:A1} because a small patch of a cell is not sufficient for the model to recognize it. In practice, these failure cases can easily be identified by deploying simple conditions searching over decreasing sizes of $s$ (which would not increase the computational cost). We note that \cref{eq:assumption_mil_equation} can also be phrased as an \texttt{OR} function across features. Intuitively, when the minimal feature size $s$ is smaller that the concept of interest, the \texttt{OR} function is no longer appropriate.

A second limitation of h-Shap pertains the way hierarchical partitions are created. We have chosen to use quadrants for their effectiveness and elegance, but this could be sub-optimal: important features may fall in-between quadrants, impacting performance. This limitation is minor, as it can be easily fixed by applying ideas of cycle spinning and averaging the resulting estimates. Furthermore, and more interestingly, hierarchical data-dependent partitions could also be employed. We regard this as future work.

\subsection{Baseline and assumptions}
Recall that all explanation methods based on feature removal--like Shapley-based explanation methods--are sensitive to the choice of baseline, i.e. the reference value used to mask features. Then, we now turn our attention to h-Shap's masking strategy, or alternatively, how to sample a reference. We recall that in this work we defined the variable $X_C$ as
\begin{equation}
    (X_C)_i =
    \begin{cases}
        X_i & \text{if}~i \in C\\
        R_i & \text{otherwise},
    \end{cases}
\end{equation}
where $R \in \mathbb{R}^{n - \lvert C \rvert}$ is a baseline value. Throughout this work, we have treated $R$ as a fixed, deterministic quantity. However, more generally, reference inputs are random variables. Let this masked input be the random variable $X_C = [\bar{X}_C, R] \in \mathbb{R}^n$, where $\bar{X}_C\in\mathbb{R}^{|C|}$ is fixed, and $R$ is a random variable. Here, we want to identify what relationships in the data distribution are important for the model, so we follow the original approach in \cite{Lundberg2017APredictions}. Indeed, the definition of Shapley values for the $i^{th}$ coefficient in \cref{eq:definition} can be made more precise by writing its expectation $\E[\f(X_{C \cup \{i\}}) - \f(X_C)]$ as
\begin{equation}
    \label{eq:conditional_masking}
    \E_{R}[\f([\bar{X}_{C \cup \{i\}}, R])~|~\bar{X}_{C \cup \{i\}}] - \E_{R}[\f([\bar{X}_{C}, R])~|~\bar{X}_C].
\end{equation}
As it can be seen, if the model $\f$ is linear, and the features are independent, then \cref{eq:conditional_masking} simplifies to
\begin{equation}
    \label{eq:unconditional_masking_diff}
    \f([\bar{X}_{C \cup \{i\}}, \E[R]]) - \f([\bar{X}_C, \E[R]]),
\end{equation}
where $\E[R]$ is an unconditional expectation which can be easily computed over the training data, and is precisely the fixed baseline we employed in this work.

How realistic are these assumptions in our case? First, the cases that we study here approximately satisfy feature independence in a local sense, and it is therefore reasonable to consider the input features as independent when $s$--the minimal feature size--is greater or similar to the size of the concept we are interested in detecting. Indeed, this is precisely true in the synthetic dataset, where each $10 \times 10$ pixels shape is sampled independently from the others. This assumption is still approximately valid in the other two experiments, where, for example, the presence or absence of a cell does not affect the content of the image so many pixels apart. On the other hand, while we have chosen very general models $\f$ which are far from linear, we argue that \cref{assump:A1} is enough to obtain a weaker sense of interpretability: looking at 
\begin{equation}
    \label{eq:unconditional_masking}
    \f([X_C, \E[R]]),
\end{equation}
and under the MIL assumption, there are only two mutually exclusive events for the subset $C$: (a)~$C$ contains at least one relevant feature, and (b)~$C$ does not contain any relevant features. When event (a) occurs, \cref{eq:unconditional_masking} will necessarily yield a high value $\approx 1$, regardless of the value of the baseline $\E[R]$. It follows that if both $C \cup \{i\}$ and $C$ contain important features, \cref{eq:unconditional_masking_diff} will be $\approx 0$; which agrees with intuition that all important features are equally important. As a result, because $\E[R]$ is fixed and \cref{assump:A1} holds, a positive value of \cref{eq:unconditional_masking_diff} is only attained if (i.e. implies that) $i$ is an important feature (and it also implies that $\E[R]$ is not important).

To summarize, the choice of using the unconditional expectation as a baseline value is approximately valid because feature independence approximately holds on a local sense, and although the models we study are highly non-linear, \cref{assump:A1} guarantees a weaker sense of interpretability. However, when these two conditions are not satisfied, one should deploy different methods to approximate the conditional distribution as in \cref{eq:conditional_masking}.
Lastly, note that our method relies on $\f$ satisfying \cref{assump:A1}, and one should wonder when this holds. Such an assumption is true when $f^*$--the true classification rule $Y=f^*(X)$--satisfies \cref{assump:A1} (which is true for a variety of problems, including the ones studied in our experiments), and $\f$ constitutes a good approximation for $f^*$. As demonstrated in this work, such assumptions are reasonable in practical settings.

\subsection{Multi-class extensions}
\begin{figure}[t]
    \centering
    \subcaptionbox{\label{fig:multiclass_original_trophozoite}$\texttt{trophozoite} = 1$.}{\includegraphics[width=0.23\linewidth]{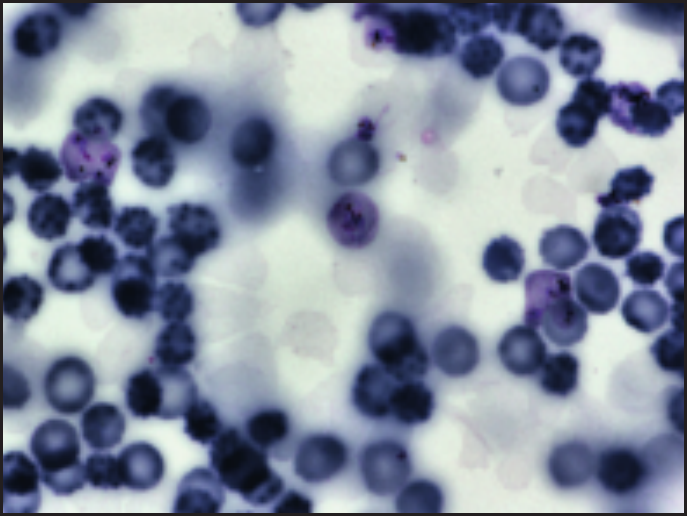}}\hfill
    \subcaptionbox{\label{fig:multiclass_explanation_trophozoite}Explanation for label \texttt{trophozoite}.}{\includegraphics[width=0.23\linewidth]{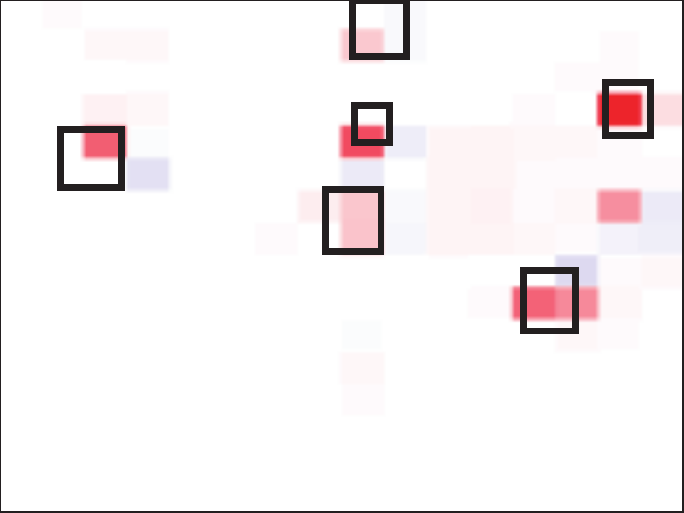}}\hfill
    \subcaptionbox{\label{fig:multiclass_original_ring}$\texttt{ring} = 1$.}{\includegraphics[width=0.23\linewidth]{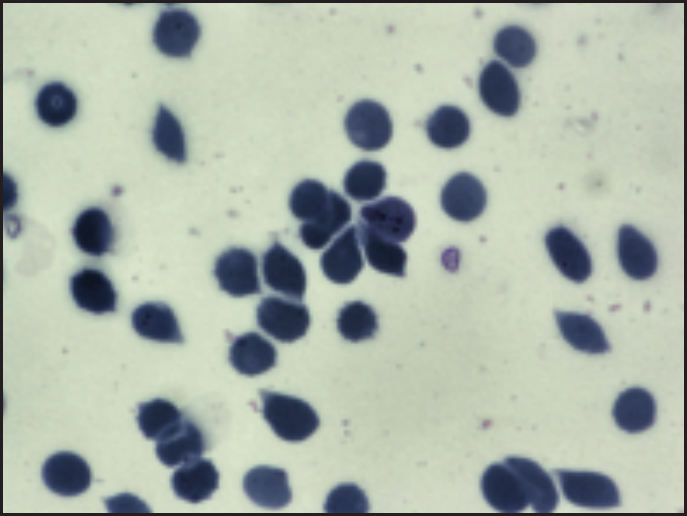}}\hfill
    \subcaptionbox{\label{fig:multiclass_explanation_ring}Explanation for label \texttt{ring}.}{\includegraphics[width=0.23\linewidth]{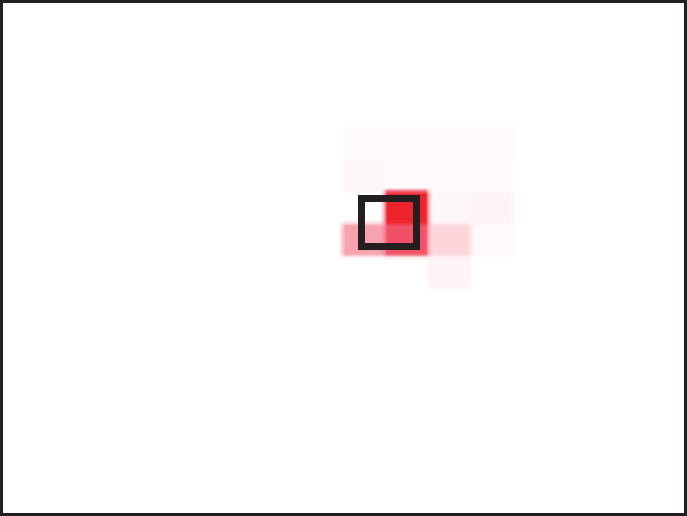}}
    \vspace{-5pt}
    \caption{\label{fig:multiclass}Example saliency maps for different labels in a multiclass setting.}
    \vspace{-15pt}
\end{figure}

Even though we have focused on binary classification tasks in this work, h-Shap can also be applied to multi-class settings. We now briefly demonstrate this by modifying the P. vivax experiment. We let $Y \in \Y = \{0, 1\}^2$, such that $Y = (\texttt{trophozoite}, \texttt{ring})$. Then, $\texttt{trophozoite} = 1$ if and only if there is at least one \emph{trophozoite} in the image, and $\texttt{ring} = 1$ if and only if there is at least one \emph{ring cell} in the image. Note that in this setting, these two classes are not mutually exclusive, as is typically the case for traditional image classifications problems. The latter setting is simply a particular case of the former. 
We randomly choose a training split that contains $80\%$ of each class, and we finetune a ResNet18 pretrained on ImageNet. We optimize all parameters for $60$ epochs with Adam \cite{Kingma2014Adam:Optimization}, using binary cross-entropy loss per class (as the classes are not mutually exclusive), learning rate of $0.0001$, weight decay of $0.00001$, and learning rate decay of $0.7$ every $7$ epochs. After training, the model achieves an accuracy of $\approx 87\%$ on each label across the held-out test set. \cref{fig:multiclass} shows saliency maps for two example images from the test set, one containing $6$ trophozoites, and one containing $1$ ring cell. h-Shap can explain every class, and it retrieves the desired, different types of cells. We regard studying the full implications and capabilities of h-shap in multi-class MIL problems as future work.

\section{\label{sec:discussion}Conclusion}

We presented a fast, scalable, and exact explanation method for image classification based on a hierarchical extension of Shapley coefficients. We showed that when the data distribution satisfies a multiple instance learning assumption, our method gains an exponential computational advantage while producing accurate--or approximate, if desired--results. Furthermore, we studied synthetic and real settings of varying complexity, demonstrating that h-Shap outperforms the current state-of-the-art methods in both accuracy and runtime, and suggesting that h-Shap acts as a weakly-supervised object detector. We have also presented and illustrated limitations of our approach, and addressing them is matter of future work. 

\bibliography{references.bib}

\newpage
\clearpage
\appendix

\setcounter{figure}{0}
\renewcommand\thefigure{\thesection.\arabic{figure}}

\section{\label{supp:algorithms}Algorithms}
\cref{alg:breadth-first} describes the breadth-first version of h-Shap presented in \cref{sec:H-SHAP}. We recall that both implementations return the set of relevant leaves $L \subseteq [n] \coloneqq \{1, \dots, n\}$ such that their Shapley coefficients are greater than a relevance tolerance $\tau$. $d$h-Shap uses an absolute tolerance, while $b$h-Shap uses a relative tolerance.

\begin{algorithm}[h]
\caption{Breadth-first h-Shap}\label{alg:breadth-first}
\begin{algorithmic}[1]
    \Procedure{$b$\text{h-Shap}}{$X, \T_0, \f$}\\
    \textbf{inputs:} image $X$, threshold $\tau\geq0$, trained model $\f$
    \State $L \gets \emptyset$
    \State $l \gets S_0$
    \While{$l$ is not empty}
        \State $\Phi_l \gets \emptyset$
        \ForAll{$S_i \in l$}
            \State $g_i \gets (X, \f, c(S_i))$
            \State $\phi_{i,1}, \dots, \phi_{i,\gamma} \gets \texttt{shap}(g_i)$
            \State $\Phi_l \gets \Phi_l \cup \phi_{i,1}, \dots, \phi_{i,\gamma}$
        \EndFor
        \State $\tau \gets \tau(\Phi_l)$
        \State $l' \gets \emptyset$
        \ForAll{$\phi_i \in \Phi_l$}
           \If{$\phi_i \geq \tau$}
                \If{$\lvert S_i \rvert \leq s$}
                    \State $L \gets L \cup S_i$
                \Else
                    \State $l' \gets l' \cup S_i$
                \EndIf
            \EndIf
        \EndFor
        \State $l \gets l'$
    \EndWhile
    \State \textbf{return} $L$
    \EndProcedure
    \State $L \gets \texttt{$b$h-Shap}(X, \T_0, \f)$
\end{algorithmic}
\end{algorithm}

\section{\label{supp:proofs}Proofs}

We summarize here the assumptions and notation used in the following results. Let $X \in \mathbb{R}^n$ be drawn so that each entry $x_i \sim a_i \mathcal{I}~+~(1 - a_i)\mathcal{I}^c$, where $a_i \sim \text{Bernoulli}(\rho)$ is a binary random variable that indicates whether the feature $x_i$ comes from an \emph{important} distribution $\mathcal{I}$, or its \emph{non-important} complement $\mathcal{I}^c$. Let
\begin{equation*}
    \f(X_C) = 1 \iff \exists i \in C:~x_i \sim \mathcal{I},~C\subseteq [n],
\end{equation*}
where $n \coloneqq \{1, \dots, n\}$ and $X_C \in \mathbb{R}^n$ is equal to $X$ in the entries of $C$ and takes value in the baseline in its complement $\bar{C}$. We denote with $\Phi_{(X, \f)} = \{\phi_1(\f), \dots, \phi_n(\f)\} \in \mathbb{R}^n$ the saliency map of $X$ where $\phi_i(\f)$ is the Shapley coefficient of $x_i$. Let $k = \|\Phi_{(X, \f)}\|_0$ be the number of reported important features by the exact Shapley coefficients. We showed earlier (see \cref{eq:exact_shap_saliency}) that under these assumptions, it follows:
\begin{equation*}
    \phi_i(\f) = 
        \begin{cases}
            1 / k & \text{if}~ x_i \sim \mathcal{I}\\
            0 & \text{otherwise}.
        \end{cases}
\end{equation*}
Furthermore, let $\T_0 = (S_0, \T_1, \dots, \T_\gamma)$ be the recursive definition of a $\gamma$-partition tree of $X$ such that $S_0 = [n]$; $\T_i, \dots, \T_\gamma$ are the subtrees branching off of $S_0$; and $c(S_i)$ are the $\gamma$ children of the node $S_i$. Recall that h-Shap explores $\T_0$ from $S_0$ and returns all relevant leaves $L \subseteq [n]$ such that their Shapley coefficient is greater than a relevance tolerance $\tau$. We denote with $\hat\T_0$ the subtree composed of the nodes visited by h-Shap, and with $\widehat{\Phi}_{(X, \f)}$ the saliency map computed by h-Shap, such that
\begin{equation*}
    \widehat{\phi}_i(\f) =
    \begin{cases}
        1/\lvert L \rvert & \text{if}~ i \in L\\
        0 & \text{otherwise}.
    \end{cases}
\end{equation*}

Now, we will provide proof of the Theorems presented in \cref{sec:H-SHAP}.

\subsection{Expected number of visited nodes, \cref{th:expectation}}
Here, we are interested in evaluating the expected number of nodes visited by h-Shap, to better characterize its computational advantage.
\begin{proof}
    \label{proof:expectation}
    Recall that $S_0$ contains all features of $X$. That is, $S_0 = [n]$. Since $x_1, \dots, x_n \in X$ are iid, so are groups of features. Then, it suffices to analyze each child of $S_0$ independently. Consider the two mutually exclusive events on the child node $c_i \in c(S_0)$: \textbf{1)} it does not contain any important features, i.e. $\nexists~j \in c_i:~\f(X_j) = 1$; and \textbf{2)} it contains at least one important feature, i.e. $\exists~j \in c_i:~\f(X_j) = 1$. Let $p_1(S_0)$ be the probability of event \num{1}, and $1-p_1(S_0)$ be the probability of event \num{2}. When event \num{2} occurs, we add one node to $\hat\T_0$, and then we explore the subtree $\hat\T_i$ branching off of $c_i$. We can recursively apply this strategy to each subtree of $\hat\T_0$, which yields
    \begin{equation}
        E[\lvert \hat\T_0 \rvert] = 1 + \gamma (1 - p_1(S_0))E[\lvert \hat\T_1 \rvert].
    \end{equation}
    We are left with evaluating $p_1(S_0)$, which simply is
    \begin{equation}
        \label{eq:p1_s0}
        p_1(S_0) = (1-\rho)^{\lvert S_0 \rvert/\gamma}
    \end{equation}
    since the probability for $x_k$ not to be important, i.e $x_k \sim \mathcal{I}^c$, is $(1 - \rho)$, and all the children $c(S_0)$ have cardinality $n/\gamma = \lvert S_0 \rvert/\gamma$ because they form a disjoint symmetric partition of $S_0$. When analyzing the $i^{th}$ subtree branching off of $S_0$, $\hat\T_i$, one has to condition on the probability of the event that $S_i$ contains at least one important feature. The probability $p'(S_i)$ of the event that $S_i$ contains at least one important feature is, again, simply $1 - (1 - \rho)^{\lvert S_i \rvert}$. Therefore
    \begin{equation}
        \label{eq:p1_si}
        p_1(S_i) = \frac{(1 - \rho)^{\lvert S_i \rvert / \gamma}(1 - (1 - \rho)^{\lvert S_i \rvert (\gamma - 1)/\gamma})}{1 - (1 - \rho)^{\lvert S_i \rvert}}
    \end{equation}
    is the conditioned probability that a child of $S_i$ does not contain any important features.
\end{proof}

\subsection{Similarity lower bound, \cref{th:cosine_similarity}}
Here, we want to find the lower bound of the similarity between $\Phi$ and $\widehat\Phi$, defined as 
\begin{equation*}
    \alpha = \frac{\langle \Phi, \widehat\Phi\rangle}{\|\Phi\|_2 \|\widehat\Phi\|_2}.
\end{equation*}

\begin{proof}
    \label{proof:cosine_similarity}
    Let $k = \| \Phi \|_0$ be the number of reported important features in $X$ as returned by the Shapley coefficients. Let $L \subseteq [n]$ be the relevant leaves returned by h-Shap. From \cref{eq:exact_shap_saliency} and \eqref{eq:Mhat} it follows that
    \begin{gather}
        \| \Phi \|_2 = \sqrt{\frac{1}{k^2}k} = \sqrt{\frac{1}{k}},\\
        \| \widehat\Phi \|_2 = \sqrt{\frac{1}{(\ell s)^2}\ell s} = \sqrt{\frac{1}{\ell s}},
    \end{gather}
    where $\lvert L \rvert = \ell s$, $\ell$ is the number of relevant leaves, and $s$ is the minimal feature size. Furthermore, we know that
    \begin{equation}
        \langle \Phi, \widehat\Phi \rangle = k\left(\frac{1}{k}\frac{1}{\ell s}\right) = \frac{1}{\ell s}.
    \end{equation}
    Therefore
    \begin{equation}
        \alpha = \frac{\langle \Phi, \widehat\Phi \rangle}{\| \Phi \|_2 \| \widehat\Phi \|_2} = \frac{\frac{1}{\ell s}}{\frac{1}{\sqrt{\ell sk}}} = \sqrt{\frac{k}{\ell s}}.
    \end{equation}
    Fixed $s$ and $k$, $\alpha$ is a monotonically decreasing function of $\ell$, which means that minimizing the similarity between $\Phi$ and $\widehat\Phi$ is equivalent to maximizing the number of leaves returned by h-Shap. When $k \leq n/s$, $\ell \leq k$, so $\alpha \geq \sqrt{k/(ks)} = 1/\sqrt{s}$. When $k > n/s$, $\lvert L \rvert = n$, therefore $\alpha \geq \sqrt{k/n}$.
\end{proof}

\section{\label{supp:partitionexplainer_comparison}Comparison with PartitionExplainer}

\begin{figure*}[t]
    \centering
    \includegraphics[width=0.53\textwidth]{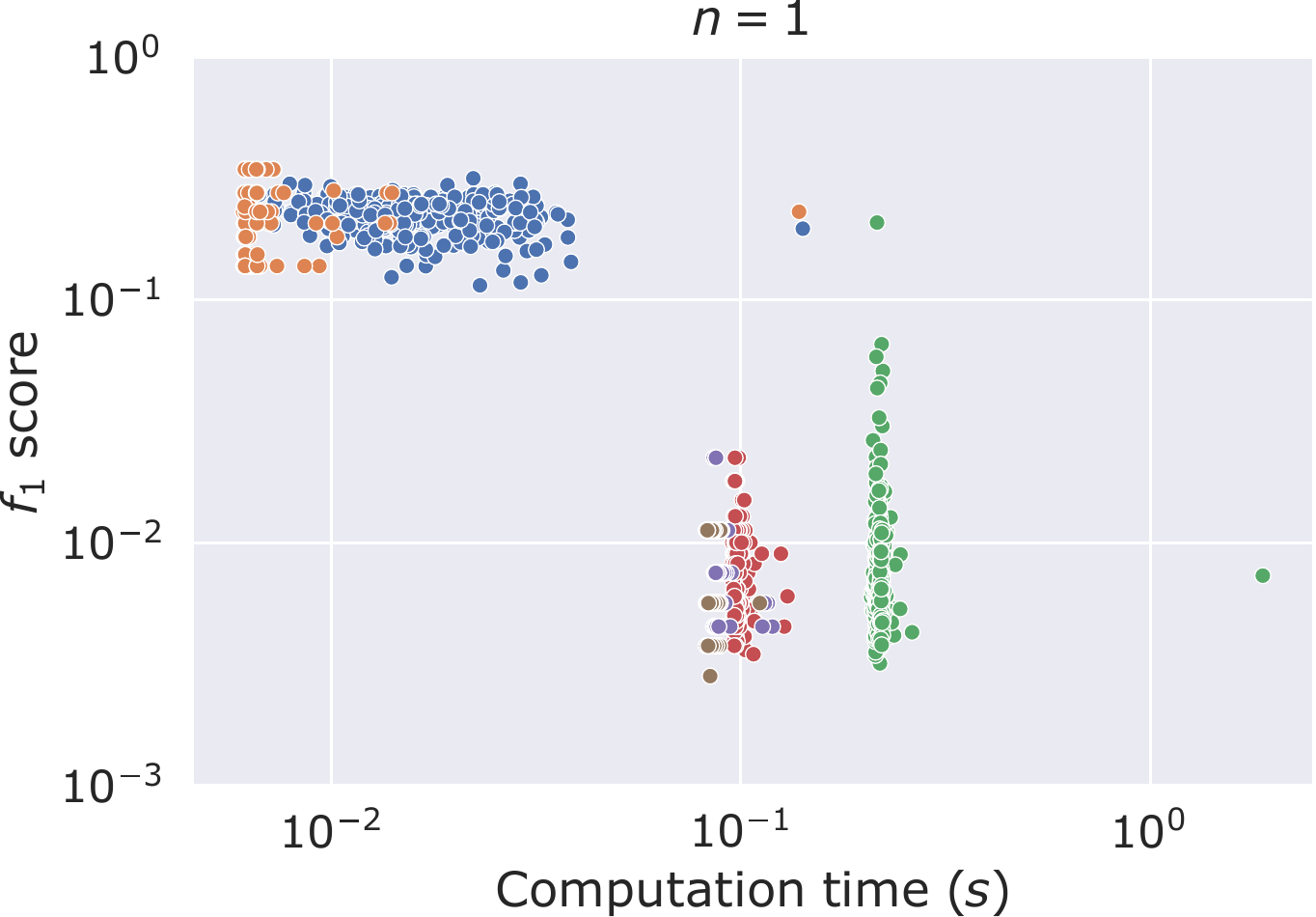}
    \includegraphics[width=0.455\textwidth]{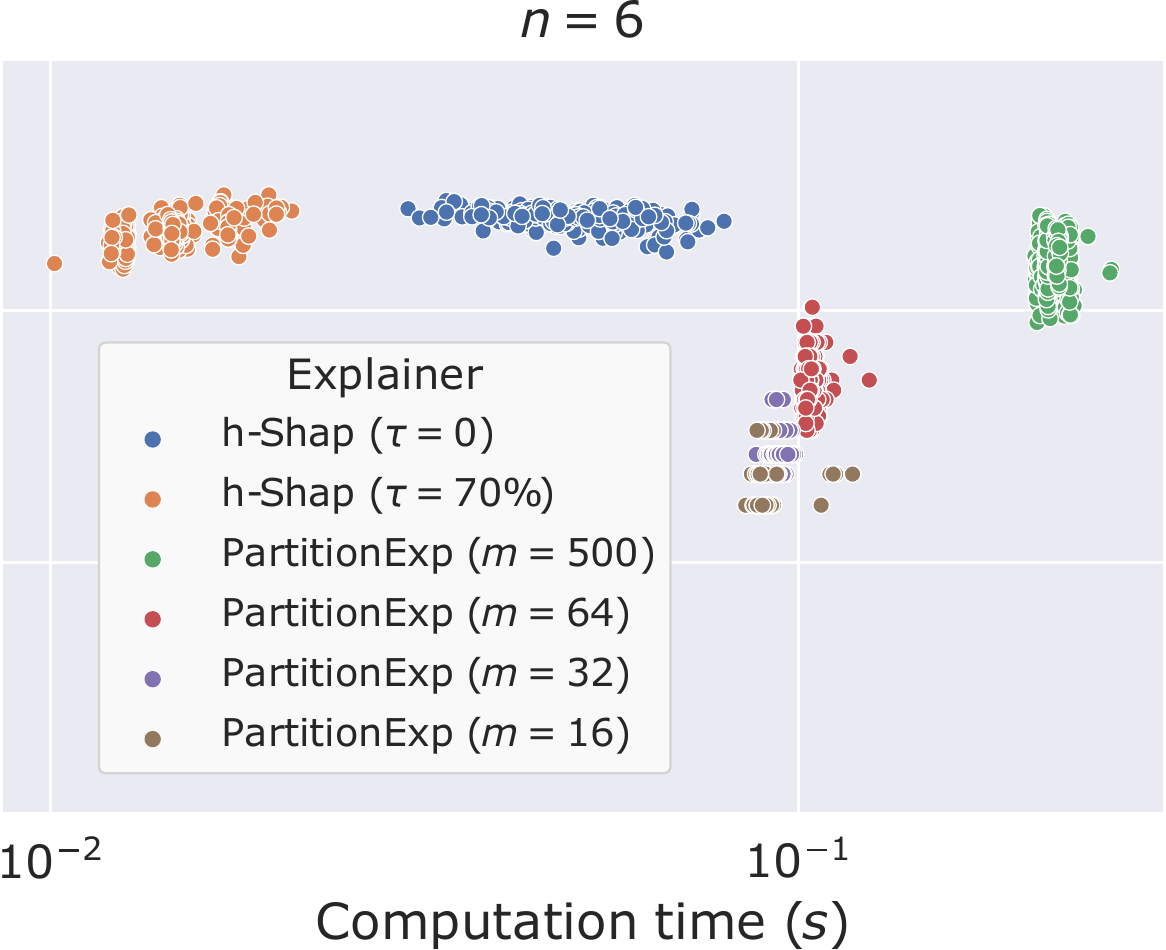}
    \caption{\label{fig:synthetic_partexp_vs_hexp}Detailed Comparison of PartitionExplainer with h-Shap in the synthetic dataset for $n=1, 6$ crosses. We use PartitionExplainer with $m = 500, 64, 32, 16$ maximal model evaluations and h-Shap with and absolute relevance tolerance of $\tau = 0$ and a relative one of $\tau = 70~\%$. $f_1$ scores are computed on binary masks obtained by thresholding the saliency maps at $1 \times 10^{-6}$ to account for noisy attributions.}
\end{figure*}

PartitionExplainer and h-Shap are closely related, as they both consider coalitions of features. The former computes Shapley coefficients recursively through a hierarchy of clusters of features, in a fashion inspired by Owen coefficients \cite{Owen1977ValuesUnions}--an extension of Shapley coefficients for games with \emph{a-priori} coalitions of players. The latter explores a quadtree of the input image, where every node corresponds to a game with $4$ players, and it only computes the exact Shapley coefficients of relevant games under a certain multiple instance learning assumption (see \cref{assump:A1}). PartitionExplainer is partition- and model-agnostic. Here, we use PartitionExplainer with axis-aligned splits, i.e. at every node, the longest axis of a partition is halved in order to generate two children nodes. That is, two iterations of PartitionExplainer produce the same leaves as one iteration of h-Shap.

Both methods reduce the exponential cost of computing Shapley coefficients. PartitionExplainer, when run on a balanced partition tree, requires quadratic runtime with respects to the number of features in the input, while h-Shap only requires a number of model evaluations that is logarithmic in the number of features.

Finally, one can control the number of clusters PartitionExplainer will explore by limiting the maximal number of model evaluations $m$. On the other hand, h-Shap provides two parameters, $s$ and $\tau$, which are informed by the problem and control the minimal feature size and relevance tolerance, respectively.
\cref{fig:synthetic_partexp_vs_hexp} showcases a detailed comparison of the two methods in the synthetic case for a different settings of their parameters.

\section{\label{supp:training_details}Experimental details}

\subsection{\label{supp:synthetic_training}Synthetic dataset}

\cref{table:synthetic_architecture} represents the network architecture used in the synthetic dataset experiment. We optimize for \num{50} epochs with Adam optimizer, learning rate $0.001$ and cross-entropy loss. 

\begin{table}[h]
\begin{center}
\begin{tabular}{|c|c|c|}
\hline
Layer     & Filter size      & Input size    \\
\hline\hline
Conv\_1    & $6 \times (3 \times 5 \times 5)$    & $3 \times 100 \times 120$ \\
ReLU\_1    & --                                  & $6 \times 96 \times 116$  \\
MaxPool\_1 & $2 \times 2$                        & $6 \times 96 \times 116$  \\
Conv\_2    & $16 \times (6 \times 4 \times 4)$   & $6 \times 48 \times 58$   \\
ReLU\_2    & --                                  & $16 \times 45 \times 55$  \\
MaxPool\_2 & $5 \times 5$                        & $16 \times 45 \times 55$  \\
FC\_1      & $1584 \times 120$                   & $1584 \times 1$      \\
ReLU\_3    & --                                  & $120 \times 1$       \\
Dropout\_1 & --                                  & $120 \times 1$       \\
FC\_2      & $120 \times 84$                     & $120 \times 1$       \\
ReLU\_4    & --                                  & $84 \times 1$        \\
Dropout\_2 & --                                  & $84 \times 1$        \\
FC\_3      & $84 \times 2$                       & $84 \times 1$        \\
\hline
\end{tabular}
\end{center}
\caption{\label{table:synthetic_architecture}Network architecture for the synthetic dataset experiment}
\end{table}

\subsection{\label{supp:malaria_lisa_training}P. vivax (malaria), LISA datasets}

In both experiments, we optimize all parameters of a pretrained ResNet18 for \num{25} epochs with Adam \cite{Kingma2014Adam:Optimization}--learning rate \num{0.0001}. We use cross-entropy loss and learning rate decay of \num{0.2} every \num{10} epochs.

\section{\label{supp:sanity_checks}Sanity checks}

\begin{figure*}[!t]
    \centering
    \includegraphics[width=0.9\linewidth]{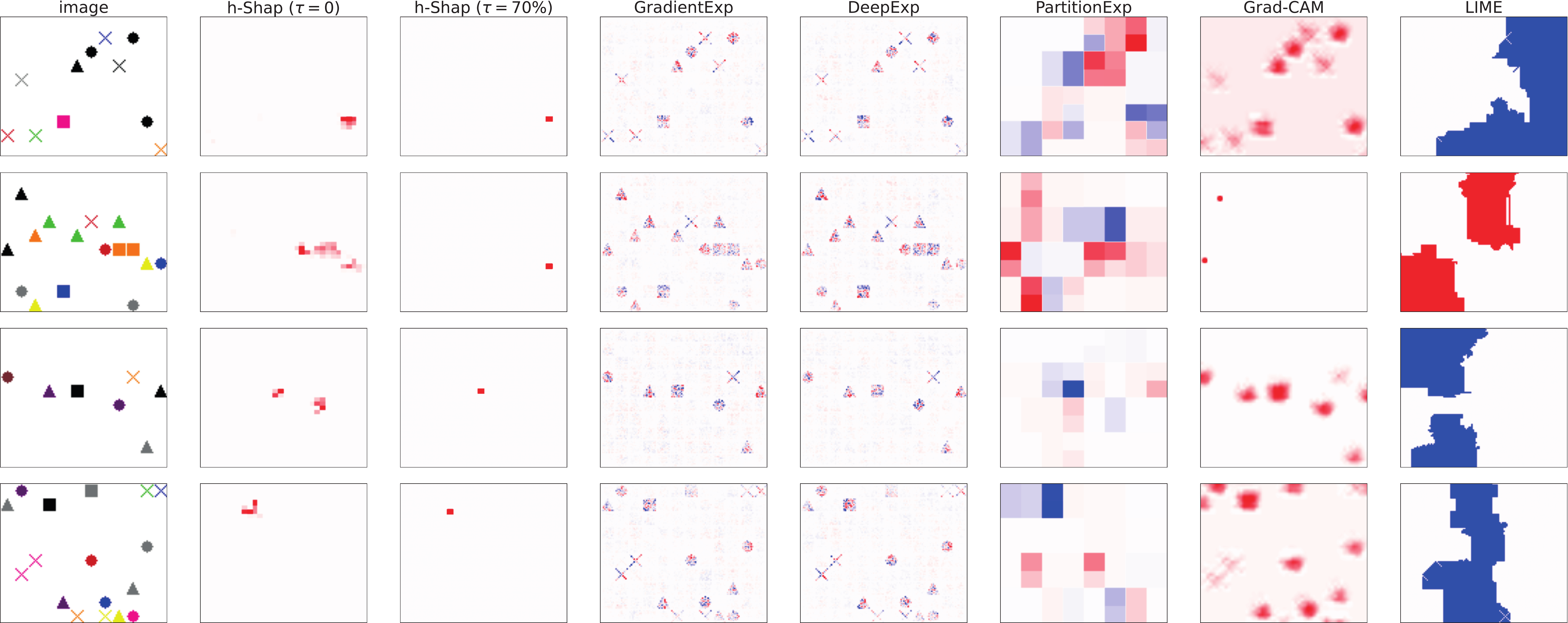}
    \caption{\label{fig:randomization_results}Examples of full model randomization tests in the synthetic dataset.}
\end{figure*}

Some interpretability methods have been shown \cite{Adebayo2018SanityMaps} to be unreliable in that they do not truly rely on what the model has learned, i.e. the precise parametrization of $\f$. For this reason, \cite{Adebayo2018SanityMaps} advocates for some \emph{sanity checks}. Following this observation, we perform full model randomization tests on all methods compared in this work. The intuition behind model randomization tests is that if the explanation method actually depends on features learned by the model, the explanations should degrade as model weights are randomized. We perform \textit{full} randomization tests in the sense that we randomly initialize \textit{all} the parameters in the simple network described above in \cref{table:synthetic_architecture}. \cref{fig:randomization_results} shows that all explanation methods employed in this work pass the model randomization test, in the sense that the saliency maps degrade completely with a random model.

\section{\label{supp:figures}Figures}
\setcounter{figure}{0}

This Appendix contains supplementary figures.

\begin{figure*}[t]
\centering
    \subcaptionbox{\label{fig:more_demos_synthetic}Synthetic dataset}
    {\includegraphics[width=0.85\textwidth]{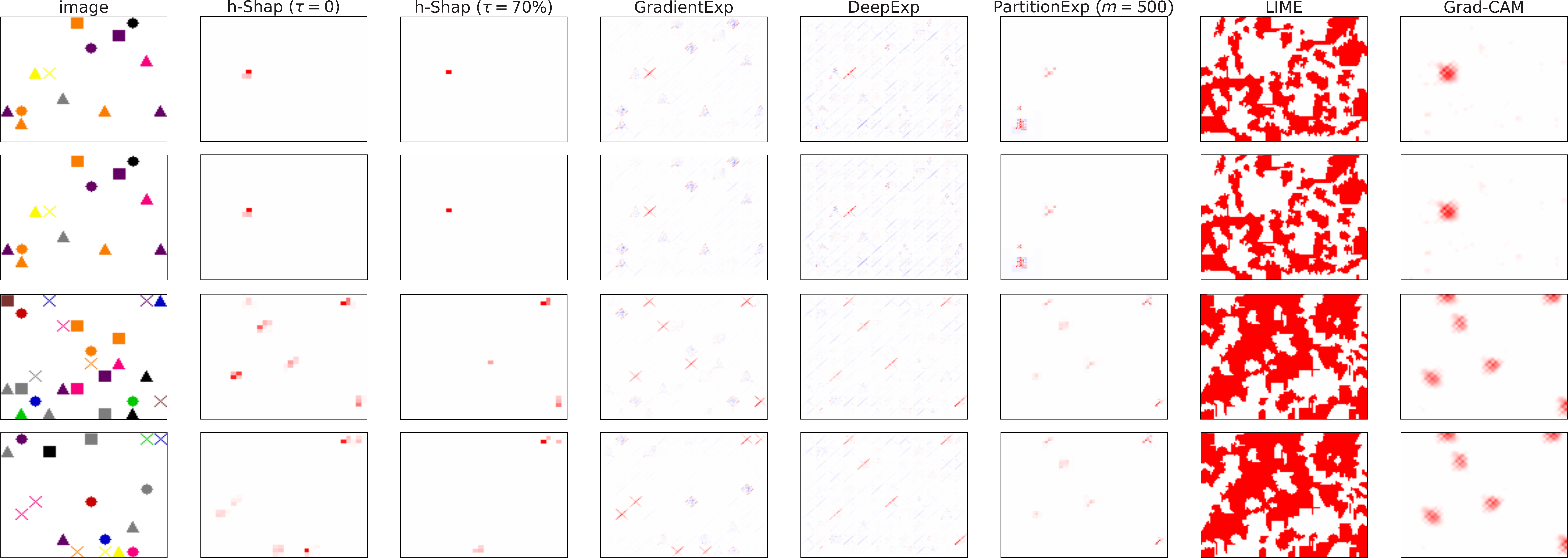}}
    \subcaptionbox{\label{fig:more_demos_malaria}BBBC041 dataset}
    {\includegraphics[width=0.85\textwidth]{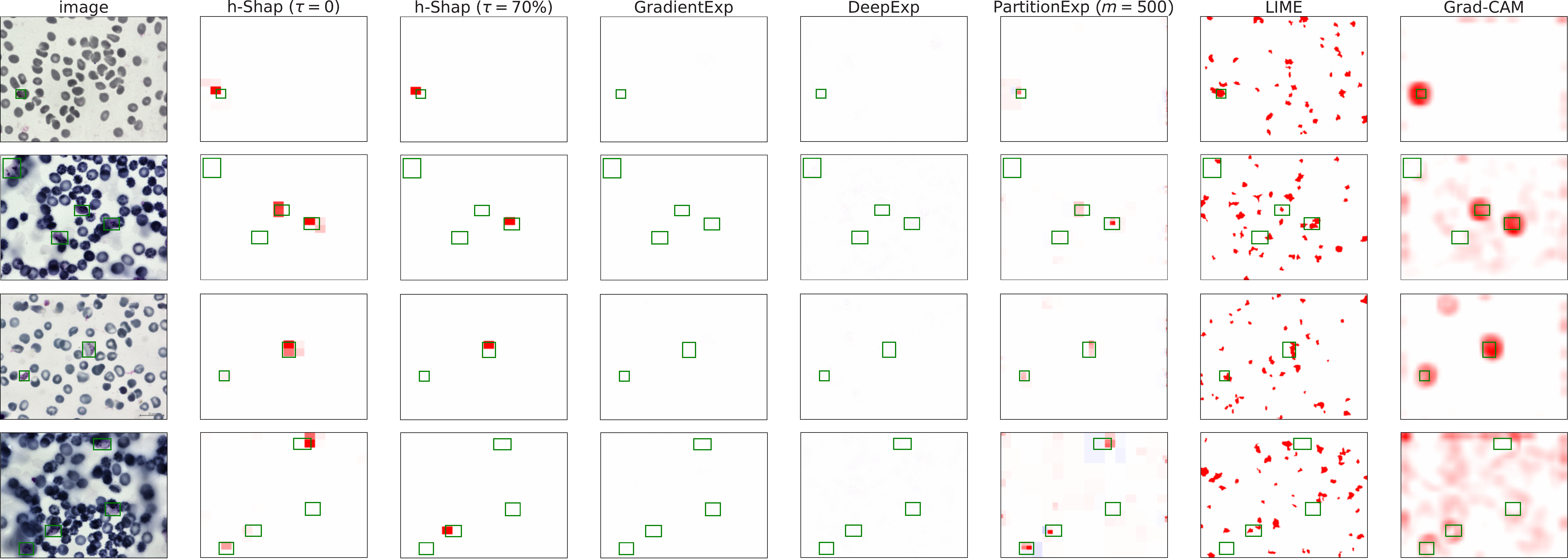}}
    \subcaptionbox{\label{fig:more_demos_lisa}LISA dataset}
    {\includegraphics[width=0.85\textwidth]{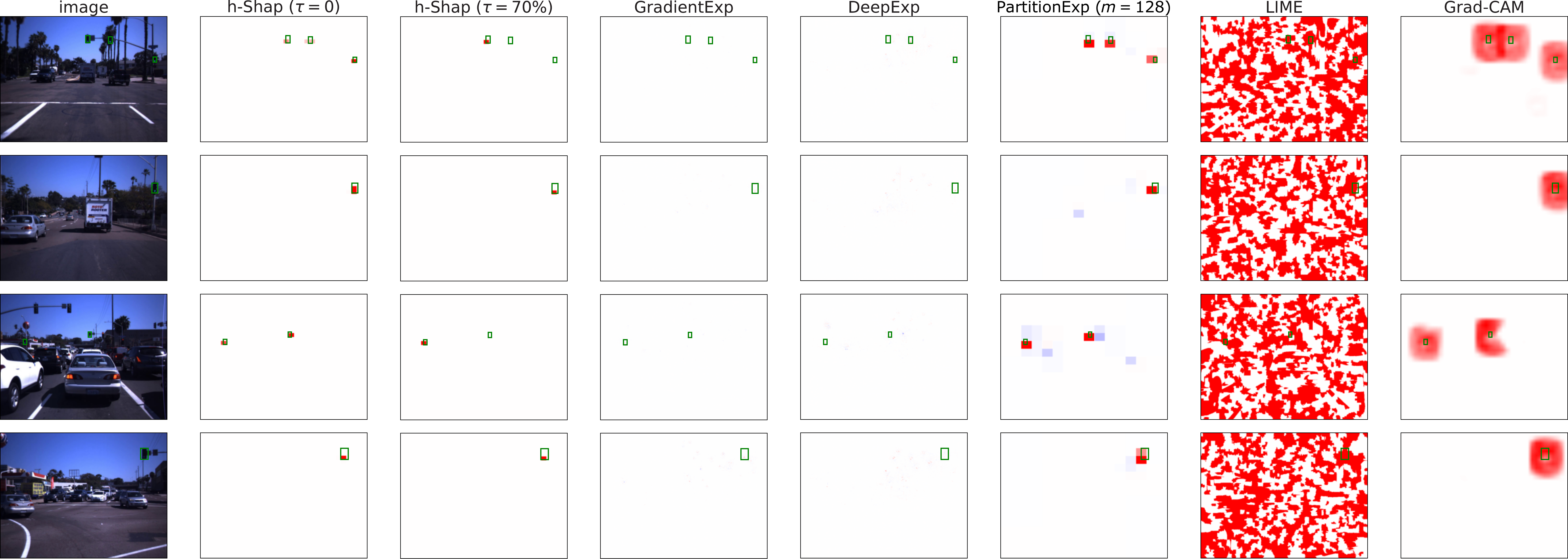}}
    \caption{\label{fig:more_demos}More examples of saliency maps.}
\end{figure*}

\begin{figure*}[t]
    \centering
    \begin{subfigure}{0.9\linewidth}
        \includegraphics[width=0.50\linewidth]{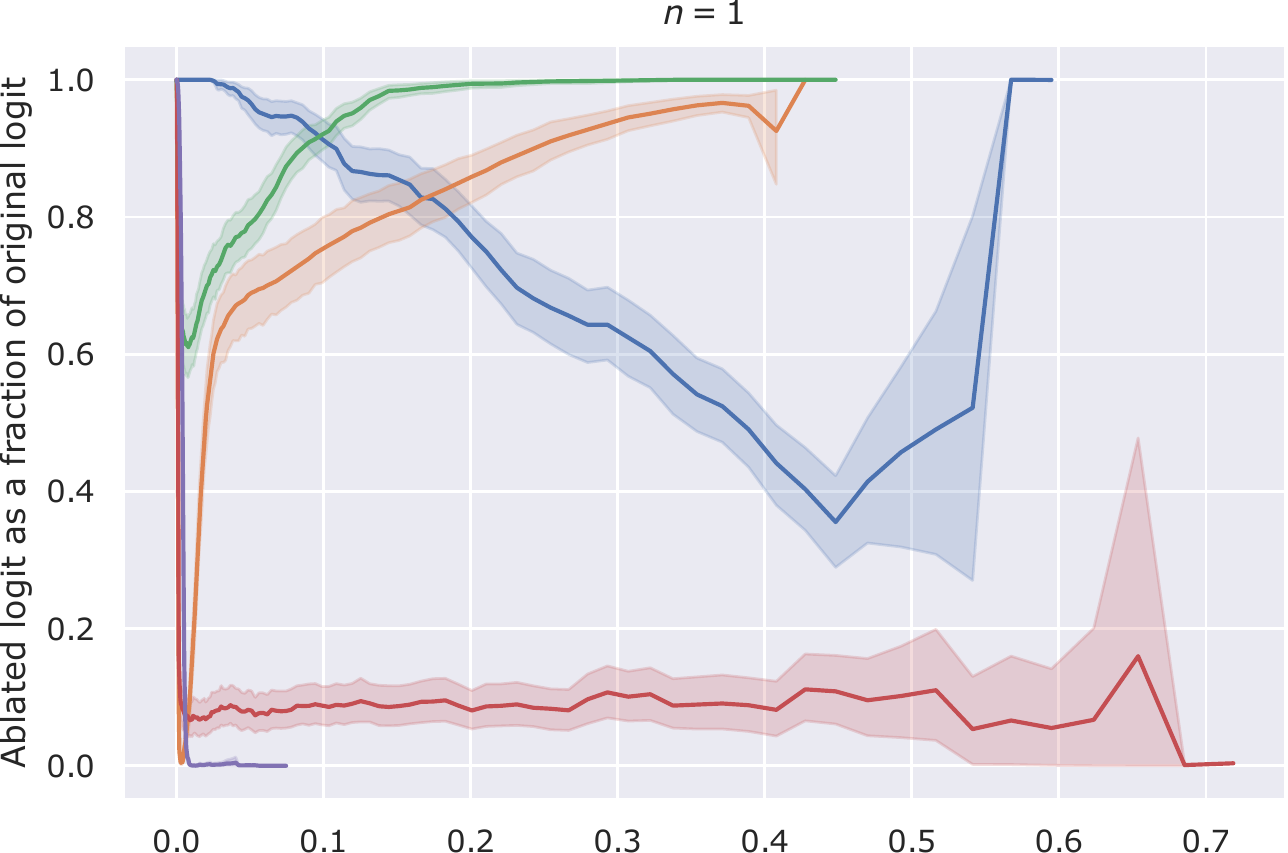}
        \includegraphics[width=0.48\linewidth]{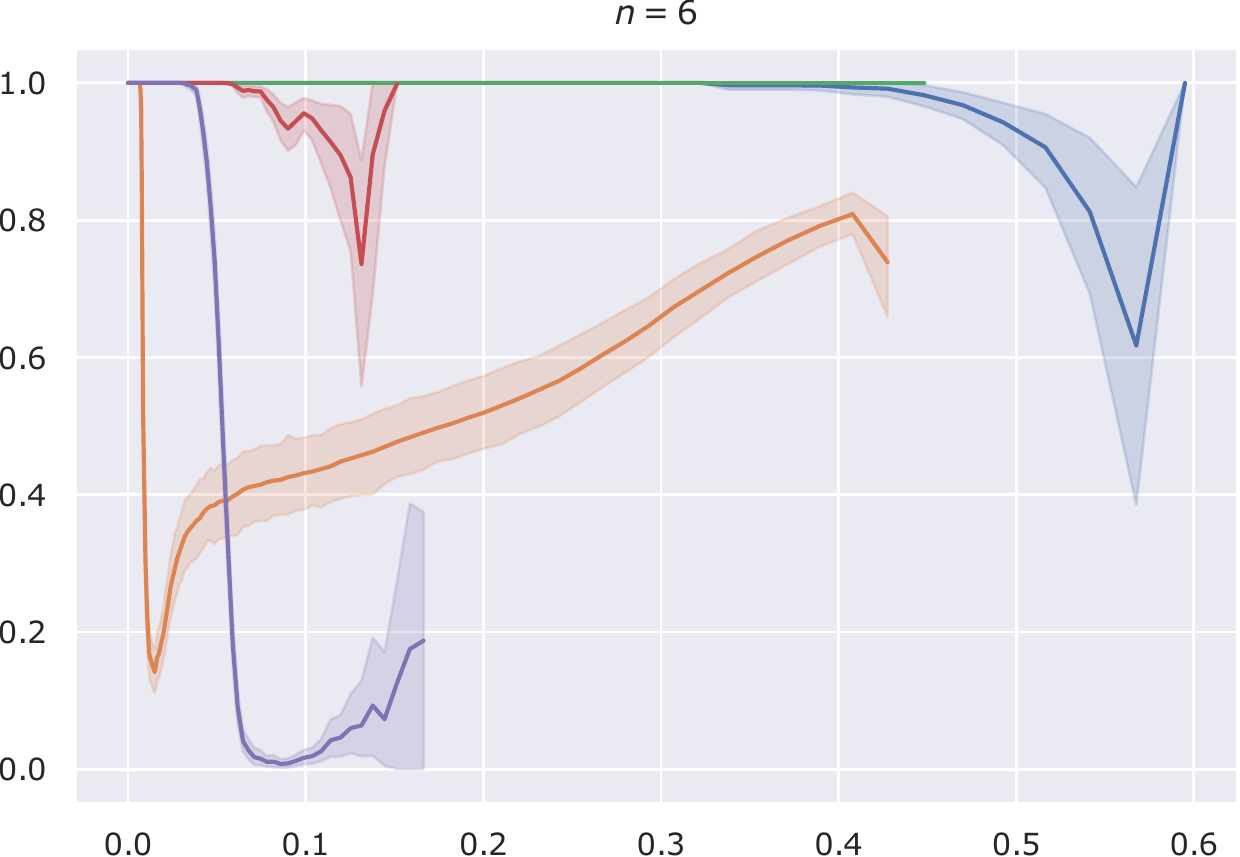}
        \caption{\label{fig:synthetic_results_ablation}Synthetic dataset. Results for $n=1, 6$ crosses.}
    \end{subfigure}
    
    \begin{subfigure}{0.9\textwidth}
        \subcaptionbox{\label{fig:malaria_results_ablation}BBBC041 dataset}
        {\includegraphics[width=0.50\linewidth]{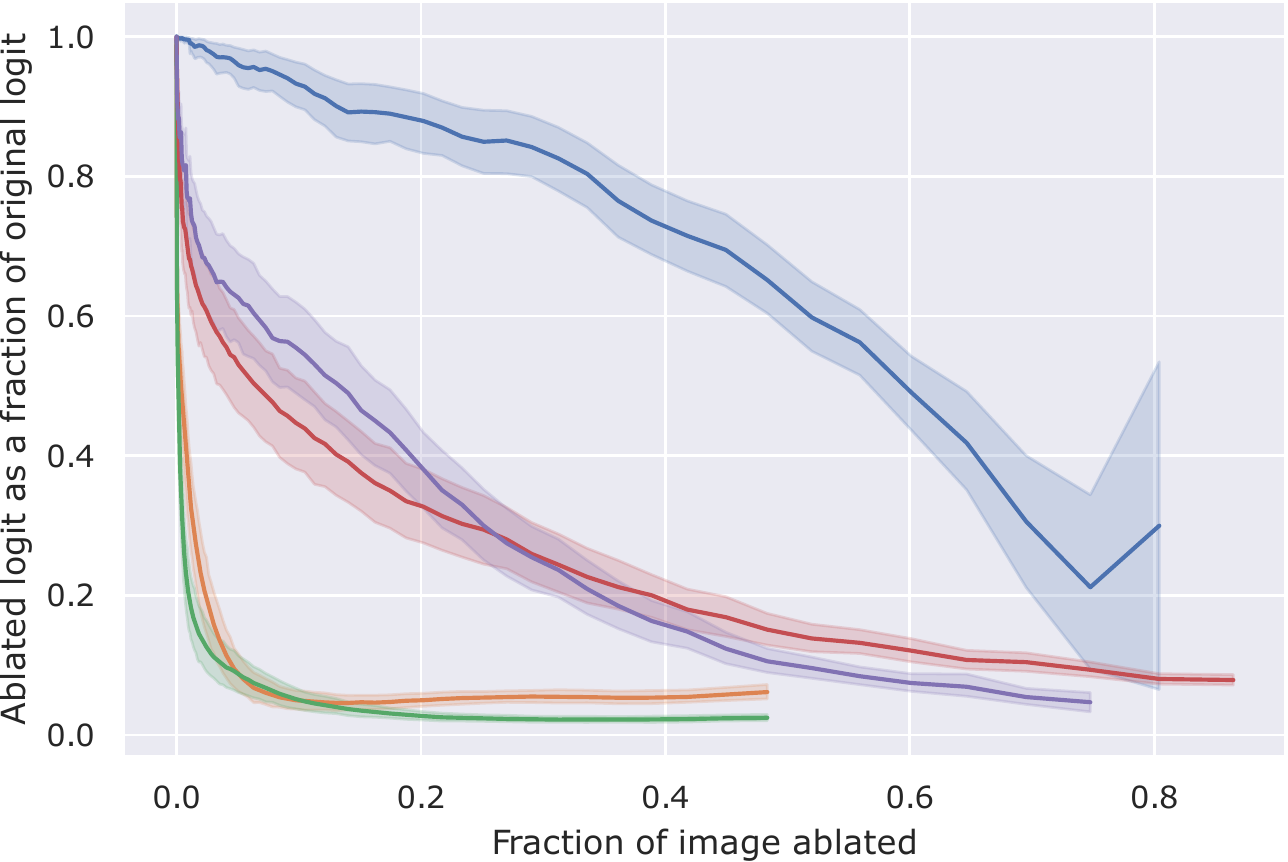}}
        \subcaptionbox{\label{fig:lisa_results_ablation}LISA dataset}
        {\includegraphics[width=0.48\linewidth]{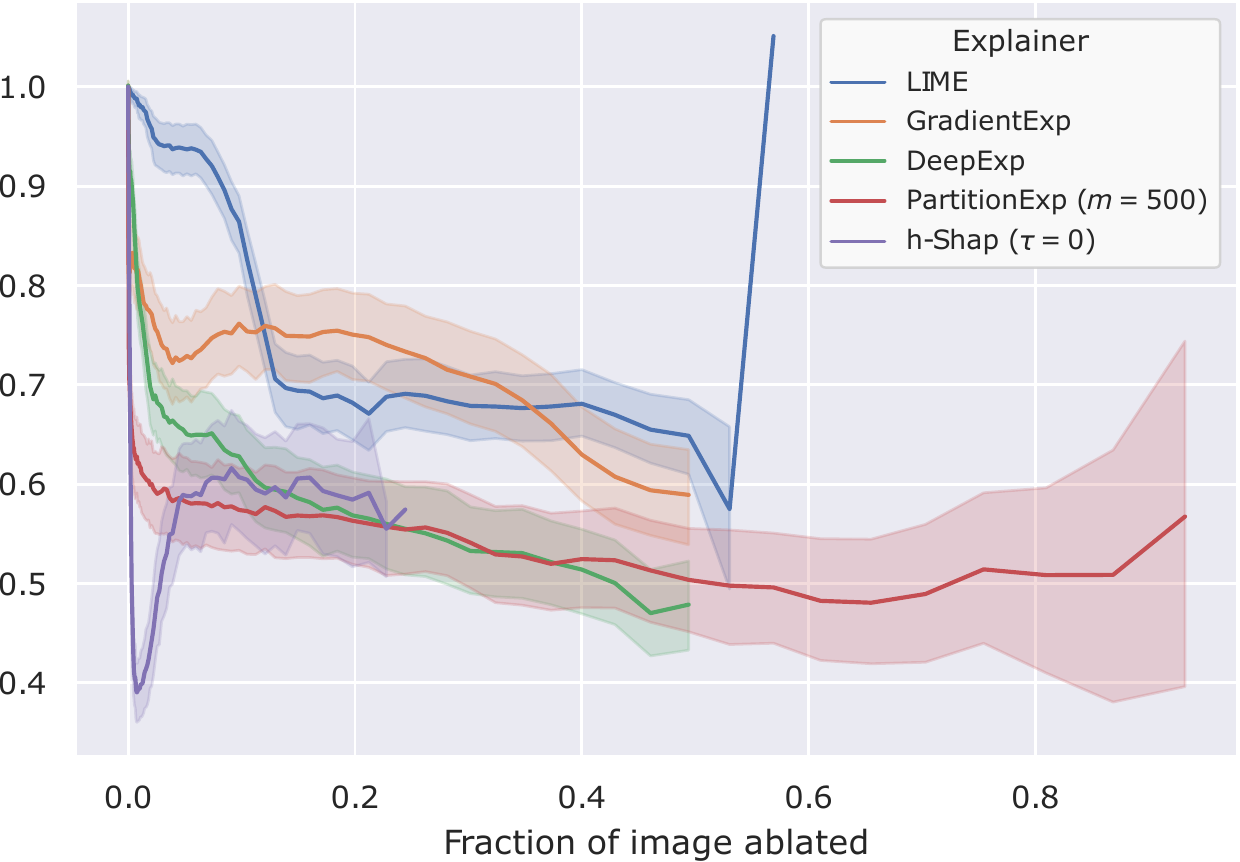}}
    \end{subfigure}
    
    \caption{\label{fig:results_ablation}Logit output compared to original logit output as a function of image ablation.}
\end{figure*}
\end{document}